\documentclass[11pt]{article}
\usepackage[table,xcdraw,dvipsnames]{xcolor}
\usepackage[]{acl}
\usepackage{times}
\usepackage{latexsym}
\usepackage[T1]{fontenc}
\usepackage[utf8]{inputenc}
\usepackage{microtype}
\usepackage{arydshln} 
\usepackage{inconsolata}
\usepackage{graphicx}
\usepackage{subcaption}
\usepackage{setspace}
\usepackage{amsmath,amsthm,amsfonts,amssymb, bm}
\usepackage{hyperref}
\usepackage{url}
\usepackage{adjustbox} 
\usepackage{listings}
\usepackage{multirow}
\usepackage{caption}
\usepackage{colortbl}
\usepackage[most]{tcolorbox}
\usepackage{titletoc}
\usepackage[capitalize]{cleveref} 
\usepackage{float}
\usepackage{tabularx}
\usepackage{booktabs}
\usepackage{array}
\usepackage{makecell}
\usepackage{enumitem}
\usepackage{bbold}
\usepackage[misc]{ifsym}
\usepackage{CJKutf8}
\usepackage{CJK}
\usepackage{color}
\usepackage{bbding}
\usepackage{longtable}

\usepackage{algorithm}
\usepackage{algorithmicx}
\usepackage{algpseudocode}

\usepackage[normalem]{ulem}
\definecolor{mygreen}{HTML}{64E8D6}
\definecolor{myblue}{HTML}{BACEFD}
\definecolor{orange}{HTML}{FFE6CC}

\definecolor{lightblue}{HTML}{DAE8FC}
\definecolor{red}{HTML}{FBBFBC}
\definecolor{green}{HTML}{D9F5D6}
\definecolor{lightgreen}{RGB}{193, 255, 193}

\usepackage{afterpage}

\usepackage{times}
\usepackage{latexsym}
\usepackage[T1]{fontenc}
\usepackage{float}
\usepackage{hyperref}
\usepackage{url}
\usepackage{booktabs}
\usepackage{xcolor}
\usepackage{arydshln}
\usepackage{longtable}
\usepackage{soul}

\usepackage{array}
\usepackage{pifont}
\usepackage{tabularx}
\usepackage{adjustbox}
\usepackage{multirow}
\usepackage{enumitem}
\usepackage{xspace}
\usepackage{tcolorbox}
\usepackage{booktabs,amsfonts,dcolumn}
\usepackage{amsmath,amsthm,amsfonts,amssymb,bm,stmaryrd,bbm}
\usepackage{colortbl}
\usepackage{wrapfig}

\usepackage{titlesec}

\titlespacing{\paragraph}{%
  0pt}{%
  0.3\baselineskip}{%
  0.5em}%
\titlespacing*{\section}{0pt}{0.4\baselineskip}{0.4\baselineskip}
\titlespacing*{\subsection}{0pt}{0.3\baselineskip}{0.3\baselineskip}

\usepackage{makecell}
\usepackage{cascadia-code}
\usepackage{courier}
\usepackage{cleveref}
\usepackage{helvet}

\definecolor{LightSteelBlue4}{RGB}{96,123,139}
\definecolor{DodgerBlue4}{RGB}{16,78,139}
\definecolor{Turquoise4}{RGB}{0,134,139}
\definecolor{Green4}{RGB}{0,139,0}
\definecolor{Brown3}{RGB}{205,85,85}
\definecolor{Azure3}{RGB}{193,205,205}

\usepackage{subcaption}
\usepackage{tikz}
\usetikzlibrary{calc}

\tikzstyle{prompt} = [rectangle,
text centered, 
minimum width = 2cm,
minimum height = 0.3cm,
font=\fontfamily{CascadiaCode-TLF}\selectfont, %
fill=LightSteelBlue4,
text=white
]

\tikzstyle{llm} = [rectangle, rounded corners,
text centered, 
draw=DodgerBlue4,
text=DodgerBlue4,
minimum width = 2cm,
minimum height = 0.6cm,
font=\small\sffamily, %
line width=1.0pt, 
fill={rgb,255:red,223;green,236;blue,248} 
]

\tikzstyle{resp} = [rectangle, %
text centered, 
draw=none,
font=\fontfamily{CascadiaCode-TLF}\selectfont,
fill=Turquoise4,
text=white
]

\tikzstyle{correct} = [rectangle, inner sep=4pt, 
text centered, 
draw=Green4,
text=Green4,
line width=1.0pt,
font=\large,
minimum width = 1.65cm,
minimum height = 0.3cm,
fill={rgb,255:red,240;green,255;blue,240}
]

\tikzstyle{wrong} = [rectangle, inner sep=4pt,
text centered, 
draw=Brown3,
text=Brown3,
line width=1.0pt,
font=\large,
minimum width = 1.65cm,
minimum height = 0.3cm,
fill={rgb,255:red,255;green,240;blue,240}
]

\tikzstyle{arrow} = [->,>=stealth,
line width=1.0pt,
draw=Azure3,
fill=Azure3
]

\tikzstyle{textlabel} = [font=\footnotesize\itshape]

\tikzstyle{sresp}=[resp,rotate=90,font=\small\fontfamily{CascadiaCode-TLF}\selectfont,inner sep=2pt]

\usepackage{listings}
\usepackage{listingsutf8} 
\usepackage[most]{tcolorbox}
\usepackage{adjustbox}
\usepackage{wrapfig}
\lstdefinelanguage{json}{alsoletter=-,}

\lstdefinestyle{agentcode}{
  inputencoding=utf8,
  basicstyle=\ttfamily\small,
  breaklines=true,
  columns=fullflexible,
  showstringspaces=false,
  backgroundcolor=\color{gray!3},
  breakautoindent=false,   
  breakindent=0pt          
}

\newtcbox{\badge}{on line, arc=3pt, colback=gray!10, colframe=gray!50,
  boxrule=0.3pt, left=4pt, right=4pt, top=1pt, bottom=1pt}

\usepackage[T1]{fontenc}
\usepackage{inconsolata}
\usepackage{upquote}

\definecolor{myPurple}{HTML}{7E007E}
\definecolor{myBlue}{HTML}{3E79CC}
\definecolor{bytedsa}{HTML}{335ab4}
\definecolor{bytedsb}{HTML}{3d8cff}
\definecolor{bytedsc}{HTML}{00c8d2}
\definecolor{bytedsd}{HTML}{79e6dd}
\definecolor{byteds1}{HTML}{a1eeec}
\definecolor{byteds2}{HTML}{e6ccff}
\definecolor{byteds3}{HTML}{ffffa5}
\definecolor{byteds4}{HTML}{00b6ff}
\definecolor{PromptBlueBack}{HTML}{EAF5FF} 
\definecolor{PromptBlueLine}{HTML}{6FA8DC} 
\definecolor{PromptGreenBack}{HTML}{EFFFF4} 
\definecolor{PromptGreenLine}{HTML}{93C47D} 
\definecolor{BoxBackNeutral}{HTML}{F8FAFC}   
\definecolor{BoxBorderNeutral}{HTML}{C9D6E5} 

\definecolor{BoxBackSoftBlue}{HTML}{F2F7FF}
\definecolor{BoxBorderSoftBlue}{HTML}{9EBCE6}

\definecolor{BoxBackSoftPurple}{HTML}{F7F3FB}
\definecolor{BoxBorderSoftPurple}{HTML}{C8B6E2}

\definecolor{BoxBackSoftTeal}{HTML}{F2FCFB}
\definecolor{BoxBorderSoftTeal}{HTML}{A6E0DD}

\newtcolorbox{observationbox}[1][]{
        colback=envfill,
        colbacktitle=envfill,
        colframe=envborder,
        arc=5pt,
        fontupper=\small,
        fonttitle=\bfseries\color{black},
        boxrule=0.5mm,
        boxsep=1mm,
        width=\linewidth,
        breakable,
        title={Observation \hfill #1},
        rounded corners,
        toptitle=0.7mm,
        bottomtitle=0.7mm
}
\newtcolorbox{goldpatchbox}[1][]{
        colback=goldpatchfill,
        colbacktitle=goldpatchfill,
        colframe=goldpatchborder,
        arc=5pt,
        fontupper=\small,
        fonttitle=\bfseries\color{black},
        boxrule=0.5mm,
        boxsep=1mm,
        width=\linewidth,
        breakable,
        title={Gold Patch \hfill #1},
        rounded corners,
        toptitle=0.7mm,
        bottomtitle=0.7mm
}
\newtcolorbox{issuebox}[1][]{
        colback=issuefill,
        colbacktitle=issuefill,
        colframe=issueborder,
        arc=5pt,
        fontupper=\small,
        fonttitle=\bfseries\color{black},
        boxrule=0.5mm,
        boxsep=1mm,
        width=\linewidth,
        breakable,
        title={Issue \hfill #1},
        rounded corners,
        toptitle=1mm
}
\newtcolorbox{agentbox}[1][]{
        colback=agentfill,
        colbacktitle=agentfill,
        colframe=agentborder,
        arc=5pt,
        fontupper=\small,
        fonttitle=\bfseries\color{black},
        boxrule=0.5mm,
        boxsep=1mm,
        width=\linewidth,
        breakable,
        title={SWE-agent \hfill #1},
        rounded corners,
        toptitle=1mm,
        lower separated=false
}
\newtcolorbox{fileviewerbox}[1]{
        enhanced,
        breakable,
        boxrule = 1.5pt,
        fontupper = \small,
        fonttitle = \bf\color{black},
        arc = 5pt,
        rounded corners,
        colframe = black,
        colbacktitle = swecream,
        colback = swecream,
        title = #1,
        left=4pt 
}
\newtcolorbox{promptbox}[1]{
    enhanced,
    breakable,
    boxrule=1pt,  
    fontupper=\small,
    fonttitle=\bfseries\color{black},
    arc=3pt,  
    rounded corners,
    colframe=black,
    colbacktitle=swecream,
    colback=swecream,
    title=#1,
    left=2mm,  
    right=2mm,  
    top=1mm,  
    bottom=1mm  
}

\tcbset{
  promptstyle/.style n args={3}{
    enhanced, breakable,
    colback=#1, colframe=#2,
    boxrule=0.6pt, arc=2mm,
    left=6pt,right=6pt, top=6pt,bottom=6pt,
    borderline west={1.6pt}{0pt}{#2},
    fonttitle=\bfseries,
    title={#3}
  }
}

\newtcolorbox{Takeaway}[1][]{
  colback=Purple!5,          
  colframe=Purple!40,        
  breakable,
  title=\textbf{Takeaway},
  fonttitle=\bfseries,
  arc=3pt,
  boxrule=0.6pt,
  left=6pt, right=6pt, top=6pt, bottom=6pt,
  enhanced,
  drop fuzzy shadow=Purple!20, 
  #1
}

\newtcolorbox{DefaultBox}[2][]{%
  colback=gray!3,          
  colframe=black!60,       
  breakable,
  title=\textbf{#2},
  fonttitle=\bfseries,
  arc=3pt,                 
  boxrule=0.6pt,
  left=6pt, right=6pt, top=3pt, bottom=3pt,
  enhanced,
  drop fuzzy shadow=gray!25,        
  #1
}

\newtcolorbox{CasePurple}[2][]{%
  colback=Purple!5,          
  colframe=Purple!40,        
  title=\textbf{#2},
  breakable,
  arc=3pt,
  boxrule=0.6pt,
  left=6pt,right=6pt,top=3pt,bottom=3pt,
  enhanced,
  drop fuzzy shadow=Purple!20, 
  #1
}

\newtcolorbox{CaseCyan}[2][]{%
  colback=cyan!5!white,
  colframe=cyan!60!black,
  coltitle=cyan!65!black,
  title=\textbf{#2},
  breakable,
  arc=3pt,
  boxrule=0.6pt,
  left=6pt,right=6pt,top=3pt,bottom=3pt,
  enhanced,
  drop fuzzy shadow=black!15,
  #1
}

\newtcolorbox{CaseBlue}[2][]{%
  colback=blue!4!white,
  colframe=blue!55!black,
  coltitle=blue!65!black,
  title=\textbf{#2},
  breakable,
  arc=3pt,
  boxrule=0.6pt,
  left=6pt,right=6pt,top=3pt,bottom=3pt,
  enhanced,
  drop fuzzy shadow=black!15,
  #1
}

\newtcolorbox{CaseGreen}[2][]{%
  colback=green!5!white,
  colframe=green!50!black,
  coltitle=green!55!black,
  title=\textbf{#2},
  breakable,
  arc=3pt,
  boxrule=0.6pt,
  left=6pt,right=6pt,top=3pt,bottom=3pt,
  enhanced,
  drop fuzzy shadow=black!15,
  #1
}

\newtcolorbox{CaseOrange}[2][]{%
  colback=orange!6!white,
  colframe=orange!65!black,
  coltitle=orange!70!black,
  title=\textbf{#2},
  breakable,
  arc=3pt,
  boxrule=0.6pt,
  left=6pt,right=6pt,top=3pt,bottom=3pt,
  enhanced,
  drop fuzzy shadow=black!15,
  #1
}

\newtcolorbox{CaseGray}[2][]{%
  colback=black!2!white,
  colframe=black!50,
  coltitle=black,
  title=\textbf{#2},
  breakable,
  arc=3pt,
  boxrule=0.6pt,
  left=6pt,right=6pt,top=3pt,bottom=3pt,
  enhanced,
  drop fuzzy shadow=black!10,
  #1
}

\newtcolorbox{AgentBoxSelf}[1][]{colback=Gray!4, colframe=Gray!45,breakable,
  title=AAA, fonttitle=\bfseries, arc=3pt, boxrule=0.6pt,
  left=6pt,right=6pt,top=6pt,bottom=6pt, enhanced, drop fuzzy shadow=Gray!25, #1}
  
\newtcolorbox{AgentBoxConsult}[1][]{colback=TealBlue!3!white, colframe=TealBlue!35!gray,breakable,
  title=AAA, fonttitle=\bfseries, arc=3pt, boxrule=0.6pt,
  left=6pt,right=6pt,top=6pt,bottom=6pt, enhanced, drop fuzzy shadow=TealBlue!18!gray, #1}
  
\newtcolorbox{AgentBoxDecomp}[1][]{colback=OliveGreen!3!white, colframe=OliveGreen!35!gray,breakable,
  title=AAA, fonttitle=\bfseries, arc=3pt, boxrule=0.6pt,
  left=6pt,right=6pt,top=6pt,bottom=6pt, enhanced, drop fuzzy shadow=OliveGreen!18!gray, #1}
  
\newtcolorbox{TrainingPrompt}[1][]{colback=TealBlue!3!white, colframe=TealBlue!35!gray,breakable,
  title=AAA, fonttitle=\bfseries, arc=3pt, boxrule=0.6pt,
  left=6pt,right=6pt,top=6pt,bottom=6pt, enhanced, drop fuzzy shadow=TealBlue!18!gray, #1}
  
\newtcolorbox{EvaluationPrompt}[1][]{colback=OliveGreen!3!white, colframe=OliveGreen!35!gray,breakable,
  title=AAA, fonttitle=\bfseries, arc=3pt, boxrule=0.6pt,
  left=6pt,right=6pt,top=6pt,bottom=6pt, enhanced, drop fuzzy shadow=OliveGreen!18!gray, #1}

\lstdefinestyle{agentcode}{
  basicstyle=\ttfamily\small,
  columns=fullflexible,
  keepspaces=true,
  showstringspaces=false,
  breaklines=true,
  breakatwhitespace=false,
  moredelim=**[s][\bfseries\color{Sepia!75!black}]{<think>}{</think>},
  moredelim=**[s][\bfseries\color{MidnightBlue!80!black}]{<agent_calls>}{</agent_calls>},
  moredelim=**[s][\bfseries\color{Orange!80!black}]{<agent_returns>}{</agent_returns>},
  commentstyle={}, 
  morecomment=[s][\bfseries\color{Sepia!75!black}]{<think>}{</think>},
  morecomment=[s][\bfseries\color{MidnightBlue!80!black}]{<agent_calls>}{</agent_calls>},
  morecomment=[s][\bfseries\color{Orange!80!black}]{<agent_returns>}{</agent_returns>},
}

\hypersetup{
    colorlinks,
    linkcolor={blue!60!black},
    citecolor={blue!60!black},
    urlcolor={blue!60!black}
}

\newcommand{\empathy}[1]{{\sethlcolor{green!50}\hl{#1}}}
\newcommand{\explore}[1]{{\sethlcolor{lightblue!60}\hl{#1}}}
\newcommand{\strategy}[1]{{\sethlcolor{yellow!20}\hl{#1}}}
\newcommand{\badresp}[1]{{\sethlcolor{red!20}\hl{#1}}}

\newcommand{\stage}[1]{\textbf{\textit{[#1]}}}

\usepackage{amsmath,amsfonts,bm}

\def\eqref#1{equation~\ref{#1}}

\def\1{\bm{1}}

\DeclareMathAlphabet{\mathsfit}{\encodingdefault}{\sfdefault}{m}{sl}
\SetMathAlphabet{\mathsfit}{bold}{\encodingdefault}{\sfdefault}{bx}{n}

\title{
\begin{center}
  \raisebox{-0.5\height}{\includegraphics[height=4.5ex]{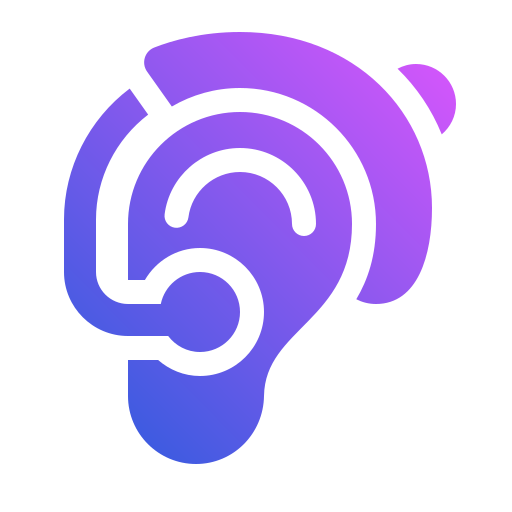}}%
  \begin{minipage}[t]{0.9\textwidth}
    \centering
    Listening to the Echo: 
    User-Reaction Aware Policy Optimization via Scalar-Verbal Hybrid Reinforcement Learning
  \end{minipage}
\end{center}
}

\author{
    Jing Ye$^{1,2}$, 
    Xinpei Zhao$^{3}$,
    Lu Xiang$^{1,2}$\Thanks{ Corresponding Author},
    Yaping Zhang$^{1,2}$,
    Chengqing Zong$^{1,2}$\\
    \footnotesize${}^1$State Key Laboratory of Multimodal Artificial Intelligence Systems, Institute of Automation, CAS, Beijing, China\\
    \footnotesize${}^2$School of Artificial Intelligence, University of Chinese Academy of Sciences, Beijing, China\\ 
    \footnotesize${}^3$Independent Researcher\\ 
    \footnotesize{yejing2022@ia.ac.cn}; \footnotesize{zhaoxinpei.zxp@alibaba-inc.com}; \footnotesize{\{lu.xiang, yaping.zhang, cqzong\}@nlpr.ia.ac.cn} \\
}

\begin{document}
\maketitle

\begin{abstract}

While current emotional support dialogue systems typically rely on expert-defined scalar rewards for alignment, these signals suffer from severe information sparsity. They cannot explain \textit{why} a response failed or \textit{how} to adapt to dynamic user states, often diverging from the actual goal of facilitating positive emotional shifts. In practice, \textbf{the most direct and reliable learning signal naturally emerges from the user's continuous reactions during the ongoing interaction}. 
Therefore, we propose \textbf{R}eaction \textbf{A}ware \textbf{P}olicy \textbf{O}ptimization (RAPO), a framework that optimizes over interaction consequences rather than rubric scores. RAPO treats dialogue as a reaction-driven process and utilizes simulated user responses to generate dense natural-language feedback through three core components: (1) Hindsight Dialogue Selection, which isolates pivotal turns that meaningfully alter user emotional trajectories; (2) Generative Hindsight Feedback, which transforms user reactions into contrastive ranking signals and natural-language critiques; and (3) Scalar–Verbal Hybrid Policy Optimization, which couples scalar reward optimization for global alignment with verbal feedback distillation for fine-grained semantic refinement. 
Extensive experiments on ESC and \textsc{Sotopia} demonstrate that RAPO significantly outperforms strong RL baselines in driving positive interaction outcomes.

\end{abstract}
\section{Introduction}
\label{sec: introduction}

Emotional Support Conversation (ESC) aims to alleviate users’ emotional distress, foster a sense of being understood, and encourage continued engagement \citep{ESConv}. 
With the rapid advancement of large language models (LLMs), recent work has made notable progress in generating fluent and seemingly empathetic responses through supervised fine-tuning (SFT) on expert demonstrations \cite{ExTES, serveforemo} or reinforcement learning (RL) \cite{zhou-etal-2023-facilitating, RLVER, Ma_2025}.
However, most existing approaches share a common optimization paradigm: they either imitate static expert-written responses or optimize scalar rewards derived from expert-defined rubrics, assuming that empathy quality can be predefined and universally agreed upon. 
This paradigm exposes a critical limitation: \textit{it treats emotional support as a static, isolated scoring task rather than a dynamic, interactive process.}

\textbf{First, optimization mismatch.} Relying on static expert rubrics creates a fundamental mismatch with actual user experiences \citep{emoharbor}. A response that scores highly on predefined empathy metrics might still feel hollow to the individual seeking support, a phenomenon we term "\textit{hallucinated empathy}". For instance, a model repeatedly outputting safe templates like “\textit{I’m sorry to hear that. That must be really hard,}” can earn perfect scalar scores while leaving the user feeling disengaged. Because static rubrics cannot dynamically observe user reactions, they optimize the model to \textit{look} supportive by expert standards rather than actually \textit{being} supportive in practice.

\textbf{Second, signal sparsity and ambiguity.} Even when scalar rewards correctly identify a poor response, compressing a complex supportive conversation into a single number leads to severe information loss \cite{FCP}. Identical low scores can mask entirely different failure modes, such as offering premature advice versus inadvertently invalidating the user's feelings. These sparse scalar signals fail to explain \textit{why} a response failed or \textit{how} the model should adapt to dynamic shifts in the user's emotional state. Without explicit natural-language guidance grounded in the conversational context, the policy lacks the necessary supervision for fine-grained semantic refinement.

\begin{figure*}
    \centering
    \includegraphics[width=1\linewidth]{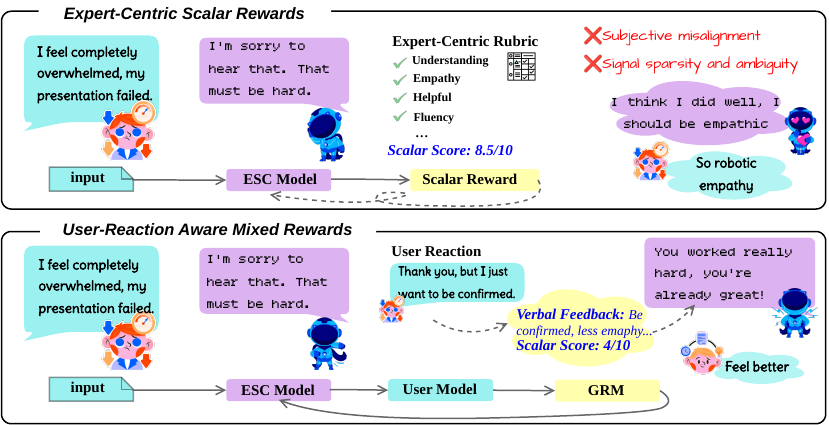}
    \caption{Comparison between expert-centric scalar rewards and user-reaction aware mixed rewards.}
    \label{fig:intro}
\end{figure*}

To bridge this gap, we claim that the most direct and reliable learning signal naturally emerges from the user’s continuous reactions during the ongoing interaction. A user's subsequent response inherently encodes whether they felt heard, validated, or alienated. \textbf{We therefore advocate a shift from rubric-matching optimization to user reaction-aware optimization.}

Building on this insight, we introduce \textbf{R}eaction \textbf{A}ware \textbf{P}olicy \textbf{O}ptimization (RAPO), a reinforcement learning framework that explicitly models user reactions through three key components:
\textbf{(1) Hindsight Dialogue Selection}, which focuses training on pivotal dialogue turns that influence the user’s emotional trajectory, avoiding overfitting to high-frequency but low-information empathy templates.
\textbf{(2) Generative Hindsight Feedback}, which employs a user simulator and a generative reward model to produce both normalized ranking scores and natural-language critiques, transforming user reactions into actionable learning signals.
\textbf{(3) Scalar–Verbal Hybrid Policy Optimization, }which couples scalar reward optimization for global alignment with verbal feedback distillation for fine-grained semantic refinement. 
Together, these components enable the model to move beyond selecting better responses toward learning to generate genuinely effective emotional support. 
Our main contributions are summarized as follows:

\begin{itemize}[itemsep= 0.05pt, topsep = 0.05pt, partopsep=0.1pt]
\item We introduce a \textbf{reaction-driven optimization paradigm} for ESC, shifting the learning objective from expert-defined rubric-matching to dynamic, interaction-driven learning.
\item We propose the \textbf{RAPO} framework that features a novel \textbf{scalar-verbal hybrid policy optimization}. By coupling ranking-based global alignment with critique-driven on-policy self-distillation, RAPO effectively resolves the signal sparsity and ambiguity inherent in standard scalar RL.
\item Extensive experiments on ESC and open-ended social environments (\textsc{Sotopia}) demonstrate that RAPO significantly outperforms strong RL baselines, proving its effectiveness and generalizability in driving positive interaction outcomes.
\end{itemize}

\section{Preliminaries}
\label{sec: preliminary}

Before presenting our method, we first review two optimization primitives closely related to our design.

\subsection{Group Relative Policy Optimization}
\label{grpo}
Group Relative Policy Optimization (GRPO) \cite{DeepSeekMath} is a variant of policy gradient methods that eliminates the need for a value function model by estimating advantages from a group of sampled outputs. Given a prompt $x$, the current policy $\pi_\theta$ generates a group of $G$ responses $\{y_i\}_{i=1}^G$, each associated with a reward $r_i$. The advantage $\hat{A}_i$ for the $i$-th response is computed by normalizing the reward within the group:
\begin{align}
\hat{A}_i = \frac{r_i - \frac{1}{G}\sum_{j=1}^G r_j}{\sigma(\{r_1, \dots, r_G\})}
\label{eq:adv}
\end{align}
where $\sigma$ denotes the standard deviation. 
The policy is then updated by maximizing the following objective:
%
\begin{equation}
\label{eq:grpo} 
\resizebox{\linewidth}{!}{$
\begin{aligned}
\mathcal{J}_{\text{GRPO}}(\theta) &= \mathbb{E}_{x, \{y_i\}} \Biggl[ \frac{1}{G} \sum_{i=1}^G  \biggl\{ \underbrace{\frac{1}{|y_i|} \sum_{t=1}^{|y_i|} L_{\text{clip}}(y_{i,t})}_{\text{Policy Loss}} - \underbrace{\beta D_{\text{KL}}(\pi_{\theta} || \pi_{\text{ref}})}_{\text{KL Penalty}} \biggr\} \Biggr] 
\end{aligned}
$}
\end{equation}
\begin{equation}
\label{eq:grpo_2}
\resizebox{\linewidth}{!}{$
\begin{aligned}
L_{\text{clip}}=&\min\left( r_{i,t}(\theta) \hat{A}_{i,t}, \right. \left.\text{clip}\left( r_{i,t}(\theta), 1 - \epsilon_{\text{low}}, 1 + \epsilon_{\text{high}} \right) \hat{A}_{i,t} \right) 
\end{aligned}
$}
\end{equation}
where $r_{i,t}(\theta) = \frac{\pi_\theta(o_{i,t} \mid q, o_{i,<t})}{\pi_{\theta_{\text{old}}}(o_{i,t} \mid q, o_{i,<t})}$ is the importance sampling ratio, $\pi_{\text{old}}$ is the policy before the update. $\epsilon$ and $\beta$ control the clipping range and the strength of the KL divergence penalty, respectively. 
This approach ensures stable updates by constraining the policy ratio while leveraging group-based advantage estimation.

\subsection{On-Policy Distillation}
\label{sdpo}
While GRPO improves stability, it relies on scalar rewards, which often fail to capture the nuanced, multi-dimensional feedback required for effective emotional support.
On-policy distillation (OPD) \citep{opd} offers a solution by aligning a student $\pi_\theta$ with a teacher $\pi_{\text{teacher}}$ through dense, token-level supervision. This is achieved by minimizing the reverse KL divergence over sampled trajectories $y \sim \pi_\theta$:
\begin{align}
\scalebox{0.95}{$
\mathcal{L}_{\mathrm{OPD}}(\theta) =
\mathbb{E}_{y \sim \pi_\theta}
\left[
\sum_{t=1}^T
\log \frac{\pi_\theta(y_t \mid x, y_{<t})}
{\pi_{\text{teacher}}(y_t \mid x, y_{<t})}
\right]
$}
\label{eq:opd}
\end{align}
From an RL perspective, the gradient $\nabla_\theta \mathcal{L}_{\mathrm{OPD}}$ can be interpreted as a policy gradient where the term $\log (\pi_{\text{teacher}}/\pi_\theta)$ acts as an implicit advantage $\hat{A}_t$. This training increases the probability that the student will generate tokens within the teacher’s distribution.

Building on this, Self-Distillation Policy Optimization (SDPO) \cite{sdpo} eliminates the external teacher by using the model's own feedback-conditioned predictions as a self-teacher:
\begin{align}
\scalebox{0.9}{$
\mathcal{L}_{\mathrm{SDPO}}(\theta) = \mathbb{E}_{y \sim \pi_\theta} \left[ \sum_{t=1}^T \log \frac{\pi_\theta(y_t \mid x, y_{<t})}{\operatorname{sg}(\pi_{\theta}(y_t \mid x, \textcolor{myPurple}{f}, y_{<t}))} \right]
$}
\label{eq:sdpo}
\end{align}
where $f$ is the environment feedback, and $\operatorname{sg}(\cdot)$ denotes the stop-gradient operation, which prevents the teacher and student distributions from co-evolving during training. By distilling feedback-informed next-token predictions back into the policy, SDPO transforms sparse outcome rewards into a dense training signal without relying on an external reward model or a separate teacher.

\section{Method}
\label{sec: method}

\begin{figure*}
    \centering
    \includegraphics[width=1\linewidth]{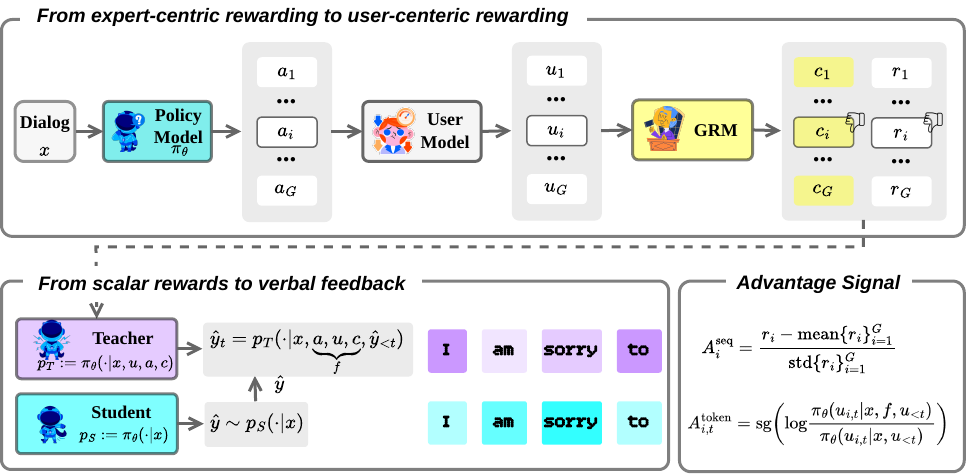}
    \caption{\textbf{Overview of the RAPO framework.} It integrates user simulation, contrastive critique generation, and scalar-verbal hybrid policy optimization.}
    \label{fig:method}
\end{figure*}

\subsection{Reaction-Aware Problem Formulation}
In conventional Reinforcement Learning from Human Feedback (RLHF), the ESC process can be modeled as follows: at each turn, given the dialogue history $x$, the policy $\pi_\theta$ generates a response $a$. The policy is then optimized to maximize the expected scalar reward $R(x, a)$: \begin{equation} 
\mathcal{J}_{\text{RLHF}}(\theta) = \max_\theta \mathbb{E}_{a \sim \pi_\theta}\left[ R(x, a) \right] 
\end{equation} 
However, as discussed in Section \ref{sec: introduction}, relying solely on a sparse scalar $R$ fails to capture the nuanced dynamics of user reactions. 

To overcome these limitations, we propose a Reaction-Aware Formulation. Instead of relying on rubric-as-reward \cite{rar}, we shift the focus to the direct outcome of the interaction: the user reaction $u \in \mathcal{U}$, where $u$ denotes the user's response to $a$. 
Our goal is to learn an optimal policy $\pi_\theta^*$ that maximizes the probability of facilitating positive user reactions. 
We define the objective as:
\begin{equation}
\mathcal{J}(\theta) = \max_\theta \mathbb{E}_{a \sim \pi_\theta, u \sim \mathcal{P}(\cdot|x, a)} \left[ \mathcal{F}(x, a, u) \right]
\end{equation}
where $\mathcal{P}(\cdot | x, a)$ is the distribution of user reactions given the context and $\mathcal{F}$ is a quality function derived from the interaction outcome $(x, a, u)$.

\subsection{Hindsight Dialogue Selection}
\label{sec: Hindsight Dialogue Selection}
Standard training paradigms typically utilize all available turns in an ESC corpus. However, a substantial fraction of these turns consist of generic empathetic templates (e.g., "\textit{I'm sorry to hear that}" or "\textit{I understand how you feel}"). 
Repeated optimization on such high-frequency but low-information patterns can lead to \emph{hallucinated empathy}, where models default to safe but vacuous responses instead of providing substantive support.

To mitigate this issue, we introduce a \textbf{Hindsight Dialogue Selection} mechanism that identifies and prioritizes \emph{pivotal turns}—interactions in which the agent's response meaningfully affects the user's emotional state or shifts the conversation's trajectory. Given a dialogue $\mathcal{D}=\{(u_j,a_j)\}_{j=1}^{N}$, we employ GPT-4o \cite{gpt4o} to perform retrospective evaluation of each turn through a hindsight-based interaction analysis. Specifically, we define a binary selection indicator for each turn:
\begin{align}
\mathbb{I}_j = \text{Hindsight-Judge}(\mathcal{D}, u_j, a_j)  
\end{align}
Model training is restricted to turns with $\mathbb{I}_j = 1$. This filtering removes redundant empathetic templates and concentrates learning on interactions that exhibit strategic support behaviors and meaningful conversational impact.

\subsection{Generative Hindsight Feedback}
\label{sec: Generative Hindsight Feedback}
To overcome the sparsity and ambiguity of scalar rewards, we propose a \textbf{Generative Hindsight Feedback} mechanism consisting of (1) group-wise exploration with user simulation and (2) contrastive critique generation.

\paragraph{Group-wise Exploration and User Simulation.}
Given a dialogue history $x$, the current policy $\pi_\theta$ samples a group of $G$ candidate responses $\{a_i\}_{i=1}^{G}$. Each candidate follows a \emph{Chain-of-Supporting} format consisting of a strategic rationale $s_i$ and the final response $y_i$, i.e., formulated as $a_i = [s_i, y_i]$.
To estimate the potential impact of these candidates, we introduce a User Simulator that provides implicit feedback. For each candidate $a_i$, the simulator generates a corresponding user reaction $u_i$ based on a predefined user persona and their current emotional state:
\vspace{-1mm}
\begin{equation}
u_i \sim \text{Simulator}(\cdot \mid x, a_i, \text{Persona})
\end{equation}
\vspace{-1mm}
This process transforms static, single-turn responses into dynamic interaction trajectories $\mathcal{T} = \{(x, a_i, u_i)\}_{i=1}^G$, providing a behavioral basis for hindsight evaluation.

\paragraph{Contrastive Critique Generation.}
By observing the entire set of simulated trajectories $\mathcal{T}$, we utilize a Generative Reward Model (GRM) \citep{liu2025inferencetimescalinggeneralistreward} to perform Contrastive Critique Generation. Unlike traditional independent scoring, the GRM processes the entire group of trajectories simultaneously. 
\begin{align}
\scalebox{0.9}{$
\{\mathcal{R}_1, \dots, \mathcal{R}_G\} = \text{GRM} \left( \cdot \mid \mathcal{T} \right)
$}
\label{eq:grm}
\end{align}
where $\mathcal{R}_i = \{\text{Rank}_i, c_i, r_i\}$, $c_i$ is the verbal hindsight critique, $r_i$ is the reward score of response $a_i$, and $\text{Rank}_i$ is the rank in the group. By forcing the GRM to rank candidates within a group, we transform the evaluation from an absolute-scoring task to a relative-preference task \citep{cpo}. This ensures that the rewards $r_i$ are inherently normalized, leading to a more stable advantage estimate $\hat{A}_i = \frac{r_i - \bar{r}}{\sigma_r}$, which could be lost under isolated scoring.

\subsection{Scalar–Verbal Hybrid Policy Optimization}
\label{sec: Dual-Level Policy Optimization}
To bridge the gap between coarse scalar rewards and fine-grained verbal feedback, RAPO adopts a hybrid optimization strategy that combines scalar reinforcement learning with verbal supervision. 
Specifically, we use GRPO to optimize the policy for ranking-based scalar rewards, while on-policy self-distillation enables the model to internalize natural-language feedback generated by the GRM.

\subsubsection{Scalar RL}
We first apply GRPO to align the policy with the ranking rewards derived in Section~\ref {sec: Generative Hindsight Feedback}. 
Given a group of sampled responses $\{a_i\}_{i=1}^G$ with contrastive ranking rewards $r_i$, the advantage $\hat{A}_i$ for each response by normalizing rewards within the group according to Eq.~\ref{eq:adv}. 
The policy $\pi_\theta$ optimized by maximizing the GRPO objective in Eq.~\ref{eq:grpo}, which encourages the model to assign higher probability to responses with better relative rankings. 
However, scalar rewards provide only coarse guidance and cannot convey detailed linguistic or strategic corrections.

\subsubsection{Verbal RL}
To internalize the rich natural-language feedback from the GRM, we introduce a verbal reinforcement learning objective via on-policy self-distillation \citep{sdpo}. The key idea is to use the policy itself as a feedback-aware teacher that corrects its own suboptimal trajectories. Rather than uniformly distilling all responses, we focus on the lowest-performing response in each group, which contains the most informative errors. By conditioning the policy on hindsight feedback describing these errors, we construct a teacher distribution that guides the original policy toward improved behavior.

\paragraph{Self-Teacher Construction.} For a given group of $G$ rollouts,  we identify the lowest-reward response $a_w = \arg \min_{a_i \in \{a_1,\dots,a_G\}} r_i $.
The teacher's input is formed by augmenting the original prompt $x$ with the corresponding verbal feedback $f_w = \{u_w, c_w\}$, where $ u_w$ denotes the user reaction to $a_w$ and $c_w$ is the natural-language critique generated by GRM that articulates the causal relationship between the response $a_w$ and the reaction $u_w$. 
Conditioned on this hindsight information and the prefix tokens of the suboptimal response $a_{w,<t}$, the policy produces a feedback-aware teacher distribution:
\begin{align}
q_\theta = \operatorname{sg}(\pi_\theta(\cdot \mid x, \textcolor{myPurple}{f_w}, a_{w,<t}))
\end{align}
Here $\operatorname{sg}(\cdot)$ denotes stop-gradient, ensuring that the teacher acts as a fixed target during distillation.

\paragraph{Optimization.}
The student policy, which only observes the original input $x$, is trained to match the teacher distribution for the suboptimal trajectory $a_w$. Following \citet{sdpo}, we employ Top-$K$ ($K=256$) approximate logit distillation to reduce computational cost while preserving gradient fidelity.

{\small
\begin{equation}
\label{eq:sdpo_loss}
\mathcal{L}_{\mathrm{SD}}(\theta) \approx
\mathbb{E}_{a \sim \pi_\theta} \sum_{t=1}^{T}
\Big(
\mathcal{L}_{\mathrm{head}}^{(t)} + \mathcal{L}_{\mathrm{tail}}^{(t)}
\Big)
\end{equation}
\vspace{-3mm}
\begin{equation}
\mathcal{L}_{\mathrm{head}}^{(t)}
=
\sum_{a_t \in \mathrm{top}K} 
\pi_\theta(a_t \mid x, a_{<t}) 
\log \frac{\pi_\theta(a_t \mid x, a_{<t})}{q_\theta(a_t)}
\end{equation}
\vspace{-3mm}
\begin{equation}
\mathcal{L}_{\mathrm{tail}}^{(t)}
=
P_{\mathrm{tail}} \log 
\frac{P_{\mathrm{tail}}}{\sum_{a_t \notin \mathrm{top}K} q_\theta(a_t)}
\end{equation}
}
where $P_{\text{tail}} = 1 - \sum_{a_t \in \text{top}_K} \pi_\theta(a_t|x, a_{<t})$ aggregates the long-tail distribution to ensure an unbiased gradient estimate. This objective encourages the student policy to reproduce the teacher’s feedback-informed token distribution.

\begin{table*}[th]
\setstretch{1}
\centering
\footnotesize
\resizebox{\textwidth}{!}{%
\begin{tabular}{lccccccc}
\toprule[1.3pt]
    \multicolumn{8}{l}{\makecell[l]{
    \textbb{PR}: Problem Resolution ~~~
    \textbb{MI}: Mood Improvement~~~
    \textbb{Div}: Diversity ~~~ 
    \textbb{HL}: Human-likeness ~~~  
    }} \\ \midrule
\multicolumn{1}{c}{} &
  \textbf{\textsc{ESConv ($\uparrow$)}} &
  \multicolumn{3}{c}{\textbf{\textsc{EmoHarbor ($\uparrow$)}}} &
  \multicolumn{3}{c}{\textbf{\textsc{Esc-Eval ($\uparrow$)}}} \\ \cmidrule(lr){2-2} \cmidrule(lr){3-5} \cmidrule(lr){6-8} 
\multicolumn{1}{c}{\multirow{-2}{*}{\textbf{Model}}} &
  \textbf{\textsc{Bert-Score}} &
  \textbf{\textbb{PR}} &
  \textbf{\textbb{MI}} &
  \textbf{\textsc{Overall}} &
  \textbf{\textbb{Div}} &
  \textbf{\textbb{HL}} &
  \textbf{\textsc{Overall}} \\ \midrule
\multicolumn{8}{l}{\textit{\textbf{Close-Source LLMs.}}}                                                                                  \\ \midrule
Doubao-Pro-32k           & \textbf{85.11} & 1.65          & 1.25          & 2.12          & \textbf{3.00} & \textbf{1.98} & \textbf{2.00} \\
GPT-4o-2024-11-20        & 84.89          & 2.56          & 2.17          & 3.20          & 2.89          & 1.36          & 1.86          \\
GPT-5-2025-08-07         & 84.26          & \textbf{3.19} & \textbf{3.06} & \textbf{3.83} & 2.78          & 1.68          & 1.98          \\ \midrule
\multicolumn{8}{l}{\textit{\textbf{Open-Source In-domain LLMs.}}}                                                                         \\ \midrule
SoulChat2.0-Qwen2-7B     & \textbf{85.75} & 1.84          & 1.20          & 2.29          & \textbf{3.00} & 2.00          & \textbf{2.00} \\
SoulChat2.0-Llama-3.1-8B & 85.65          & 1.88          & 1.27          & 2.37          & \textbf{3.00} & \textbf{2.01} & \textbf{2.00} \\
PsyChat-Qwen2.5-7B       & 84.99          & \textbf{2.03} & \textbf{1.54} & \textbf{2.55} & 2.99          & 1.60          & 1.96          \\ \midrule
\multicolumn{8}{l}{\textit{\textbf{Trained LLMs on ESConv Dataset.}}}                                                                     \\ \midrule
\rowcolor[HTML]{ECE2FE} 
Qwen-2.5-7b              & 85.67          & 1.69          & 1.21          & 2.10          & 2.61          & 0.96          & 1.27          \\
$+$ SFT-Full             & \textbf{85.85} & 1.13          & 1.00          & 1.58          & 2.89          & \textbf{2.02} & 1.97          \\
$+$ SFT-HDS              & 85.61          & 1.22          & 1.06          & 1.72          & 2.97          & 1.98          & \textbf{2.00} \\
$+$ GRPO                 & 85.57          & 1.60          & 1.04          & 2.02          & 2.99          & 1.66          & 1.98          \\
$+$ CPO                  & 85.60          & 1.76          & 1.25          & 2.15          & 2.98          & 1.83          & 1.99          \\ \hdashline
$+$ RAPO                 & 85.65          & \textbf{1.86} & \textbf{1.26} & \textbf{2.27} & \textbf{3.00} & 1.84          & \textbf{2.00} \\ \midrule
\rowcolor[HTML]{ECE2FE} 
Qwen-3-8b                & 84.42          & 1.89          & 1.62          & 2.59          & \textbf{2.91} & 1.27          & 1.93          \\
$+$ SFT-Full             & \textbf{85.89} & 1.09          & 1.00          & 1.54          & 2.14          & 2.01          & 1.94          \\
$+$ SFT-HDS              & 85.80          & 1.29          & 1.10          & 1.72          & 2.22          & \textbf{2.02} & 1.97          \\
$+$ GRPO                 & 84.48          & 1.99          & 1.57          & 2.58          & 2.89          & 1.76          & 1.89          \\
$+$ CPO                  & 85.01          & 1.99          & 1.67          & 2.58          & 2.93          & 1.87          & 1.99          \\ \hdashline
$+$ RAPO                 & 85.56          & \textbf{2.03} & \textbf{1.73} & \textbf{2.67} & \textbf{2.91} & 2.01          & \textbf{2.00} \\ 
\midrule
\rowcolor[HTML]{ECE2FE} 
Llama-3-8b               & 85.47          & 2.03          & 2.00          & 2.99          & 2.90          & 1.64          & 1.97          \\
$+$ SFT-Full             & \textbf{85.84}          & 1.25          & 1.00          & 1.66          & 2.91          & \textbf{2.01}          & \textbf{2.00}          \\
$+$ SFT-HDS              & \textbf{85.84}          & 1.20          & 1.07          & 1.69          & 2.93          & \textbf{2.01}          & \textbf{2.00}          \\
$+$ GRPO                 & 85.46          & 2.03          & 1.56          & 2.59          & 2.92          & 1.69          & 1.99          \\
$+$ CPO                  & 85.33          & 2.40          & 2.03          & 3.03          & 2.96          & 2.00          & 1.99          \\
$+$ RAPO                 &85.56           & \textbf{2.46}          & \textbf{2.05}          & \textbf{3.04 }         & \textbf{3.00}          & 1.94         &  \textbf{2.00}        \\ \bottomrule[1.3pt]
\end{tabular}%
}
\caption{\textbf{Performance on Emotional Support Conversation Benchmarks.} ``HDS'' denotes the Hindsight Dialogue Selection strategy described in Sec.~\ref{sec: Hindsight Dialogue Selection}, while ``Full'' indicates training on the complete dataset. Note that evaluation scales differ across benchmarks: \textbf{EmoHarbor} uses a five-point Likert scale (1–5), whereas \textbf{ESC-Eval} uses a five-point Likert scale (0–4). Detailed results for each evaluation dimension are provided in Tables~\ref{tab:main_ESC}, \ref{tab:main_emoharbor}, and \ref{tab:main_esc_eval}.}
\label{tab:ESC}
\end{table*}

\subsubsection{Final Objective}
The overall RAPO objective integrates scalar alignment with linguistic refinement:
\begin{equation}
\mathcal{J}(\theta) = \mathcal{J}_{\text{GRPO}}(\theta) - \eta \mathcal{L}_{\text{SD}}(\theta)
\label{eq: final loss}
\end{equation}
where $\eta$ controls the strength of the verbal supervision signal. By jointly optimizing these objectives, RAPO balances high-level strategy selection guided by scalar rewards with fine-grained behavioral correction induced by natural-language feedback. More theoretical analysis and discussion are provided in Appendix \ref{app: derivation} and \ref{app: discussion}, respectively.

\section{Experimental Results}
\label{results}

We present experiments to (i) quantify improvements over standard RL baselines and (ii) identify key design choices. Detailed settings, implementation details, and additional results are provided in Appendix~\ref{app: Evaluation Details and Setup}, ~\ref{app: Implementation Details}, and ~\ref{app:experiments}.

\subsection{Is reaction-aware optimization effective for Emotional Support Agents?}
\label{sec:main_results_esc}

We train ESC agents on ESConv \citep{ESConv} and evaluate on the ESConv test set and two interactive benchmarks, EmoHarbor \citep{emoharbor} and ESC-Eval \citep{ESC-Eval}. Table~\ref{tab:ESC} summarizes the results.

\noindent \textbf{(1) RAPO consistently demonstrates superior performance over other RL baselines on both EmoHarbor and ESC-Eval.} Specifically, Qwen-2.5-RAPO improves over GRPO by 12.4\% on EmoHarbor. Compared with CPO, it further improves EmoHarbor by 5.6\%.

\noindent \textbf{(2) The performance gains provided by RAPO are less pronounced on reasoning-focused models compared to instruction-tuned models.} While Qwen-3-RAPO still achieves measurable improvements on EmoHarbor (+3.5\% vs. GRPO and +3.5\% vs. CPO) and ESC-Eval (+5.8\% vs. GRPO and +0.5\% vs. CPO), the improvement margins are noticeably smaller than those observed with Qwen-2.5-RAPO. This suggests that models with stronger inherent reasoning capabilities already exhibit a higher baseline proficiency, leaving relatively less headroom for RL-based alignment.

\noindent  \textbf{(3) RAPO maintains strong semantic consistency with human reference responses.} On the ESConv, Qwen-2.5-RAPO and Qwen-3-RAPO achieve high BERT-Score values of 85.65 and 85.56, respectively. This indicates that the RAPO effectively optimizes support strategies without degrading the fundamental linguistic quality or deviating from appropriate conversational norms.

\noindent \textbf{(4) Utilizing a curated subset of high-quality data is sufficient for optimal performance.}
We compare the results of SFT trained on the full dataset with those trained on data selected using Hindsight Dialogue Selection, denoted as SFT-Full and SFT-HDS, respectively. The experimental results show that SFT-HDS achieves performance comparable to SFT-Full on ESConv, while demonstrating significantly better stability and generalization on EmoHarbor and ESC-Eval. This finding suggests that simply scaling up the volume of SFT data does not necessarily lead to improved emotional support capabilities. Instead, training should prioritize critical dialogue turns that require complex emotional reasoning.

\subsection{Can RAPO Generalize to Other Conversational Scenarios?}
\label{sec:main_results_sotopia}
\vspace{-2mm}
\begin{table}[th]
\setstretch{1.2}
\centering
\resizebox{\columnwidth}{!}{%
\begin{tabular}{lcccc}
\toprule[1.3pt]
\multicolumn{1}{c}{}                                  & \multicolumn{2}{c}{\textbf{\textsc{Sotopia-All} ($\uparrow$)}} & \multicolumn{2}{c}{\textbf{\textsc{Sotopia-Hard} ($\uparrow$)}} \\ \cmidrule(lr){2-3} \cmidrule(lr){4-5} 
\multicolumn{1}{c}{\multirow{-2}{*}{\textbf{Method}}} & \textsc{Goal} & \textsc{Overall} & \textsc{Goal}  & \textsc{Overall} \\ \midrule
GPT-4o              & 8.19            & \textbf{3.76}  & 6.97            & \textbf{3.46}   \\
GPT-5$^{*}$         & \textbf{8.95}   & —              & \textbf{7.21}   & —               \\
Doubao-Pro-32k      &8.12             & 3.70           & 6.49            & 3.22           \\
DeepSeek-R1         & 7.97            & 3.40           & 5.86            & 2.73            \\
Claude-3.5-Sonnet   & 8.29            & 3.71           & 6.33            & 3.09            \\
\midrule
\midrule
\rowcolor[HTML]{ECE2FE} 
Qwen2.5-7B-Instruct         & 7.91          & 3.55          & 6.21          & 3.01         \\ 
$+$ PPO$^{*}$            & 8.25            & —              & 6.63            & —               \\
$+$ GRPO$^{*}$           & 8.31            & —              & 6.97            & —               \\
$+$ RLOO$^{*}$           & 7.13            & —              & 5.70            & —               \\
$+$ GRPO iStar$^{*}$     & \textbf{8.42}   & —              & 7.11            & —               \\
$+$ CPO               & 8.13   & 3.77              & 7.14            & 3.48               \\
\midrule
$+$ RAPO            & 8.41   & \textbf{3.87}  & \textbf{7.19}   & \textbf{3.54}   \\ \bottomrule[1.3pt]
\end{tabular}%
}
\caption{\textbf{Performance on \textsc{Sotopia}.} The evaluated model interacts with itself. 
“Hard” denotes challenging scenarios requiring advanced reasoning; “All” denotes the full set. 
\textsc{GPT-4o} serves as a proxy for human judgment. Metrics include GOAL (0–10 goal completion) and OVERALL (mean of 7 dimensions), averaged over 3 runs. Entries marked * are from \citet{liu2026agentic}.}
\label{tab:sotopia}
\end{table}

To evaluate the generalizability of our approach beyond emotional support, we benchmark RAPO on \textsc{Sotopia}, a multi-agent social intelligence framework that requires complex social negotiation and interaction. The results are summarized in Table~\ref{tab:sotopia}.

\textbf{(1) RAPO outperforms established RL baselines, including PPO, GRPO, and RLOO.} Across all tasks, RAPO achieves an overall average score of 3.87 and a goal completion score of 8.41, demonstrating its robustness in managing multifaceted social dynamics. 
\textbf{(2) RAPO exhibits exceptional performance in high-difficulty scenarios.} Notably, on the \textsc{Sotopia-Hard} subset, RAPO improves the goal completion score by 3.2\% (6.97 $\rightarrow$ 7.19) compared to GPT-4o, suggesting that RAPO is particularly effective when the social objectives are complex.
\textbf{Overall, these results confirm that RAPO’s utility extends beyond emotional support.} By effectively modeling participant reactions as feedback signals and optimizing with a combination of scalar and verbal feedback, RAPO enhances general social intelligence and goal-oriented reasoning, demonstrating its effectiveness across a wide range of dialogue-based applications.

\subsection{Ablation Study}
\label{sec:ablation}

\begin{table*}[th]
\centering
\setstretch{1.2}
\resizebox{\textwidth}{!}{%
\begin{tabular}{lccccccccccc}
\toprule[1.3pt]
\multicolumn{1}{c}{} &
  \textbf{\textsc{ESConv}} &
  \multicolumn{3}{c}{\textbf{\textsc{EmoHarbor}}} &
  \multicolumn{3}{c}{\textbf{\textsc{Esc-Eval}}} &
  \multicolumn{2}{c}{\textbf{\textsc{Sotopia-All}}} &
  \multicolumn{2}{c}{\textbf{\textsc{Sotopia-Hard}}} \\ \cmidrule(lr){2-2} \cmidrule(lr){3-5} \cmidrule(lr){6-8} \cmidrule(lr){9-10} \cmidrule(lr){11-12} 
\multicolumn{1}{c}{\multirow{-2}{*}{\textbf{Model}}} &
  \textsc{Bert-Score} &
  \textbb{PR} &
  \textbb{MI} &
  \textsc{Overall} &
  \textbb{Div} &
  \textbb{HL} &
  \textsc{Overall} &
  \textsc{Goal} &
  \textsc{Overall} &
  \textsc{Goal} &
  \textsc{Overall} \\ \midrule
\rowcolor[HTML]{ECE2FE} 
RAPO &
  \textbf{85.65} &
  1.86 &
  1.26 &
  \textbf{2.27} &
  \textbf{3.00} &
  1.84 &
  \textbf{2.00} &
  \textbf{8.41} &
  \textbf{3.87} &
  7.19 &
  \textbf{3.54} \\ \midrule
w/o URM &
  85.33 &
  \textbf{1.90} &
  \textbf{1.28} &
  2.18 &
  \textbf{3.00} &
  \textbf{1.85} &
  \textbf{2.00} &
  8.17 &
  3.79 &
  \textbf{7.23} &
  3.45 \\
w/o SD &
  85.54 &
  1.87 &
  1.24 &
  2.22 &
  \textbf{3.00} &
  1.73 &
  \textbf{2.00} &
  8.08 &
  3.70 &
  7.13 &
  3.38 \\
w/o URM + SD &
  85.60 &
  1.76 &
  1.25 &
  2.15 &
  2.98 &
  1.83 &
  1.99 &
  8.13 &
  3.77 &
  7.14 &
  3.48 \\ \bottomrule[1.3pt]
\end{tabular}%
}
\caption{\textbf{Ablation results.} \textbf{RAPO w/o URM} replaces user-centric reactions with expert-defined rubric rewards; \textbf{RAPO w/o SD} removes the verbal feedback-based self-distillation mechanism; \textbf{RAPO w/o URM+SD} reduces the framework to a conventional rubric-as-reward RL setup.
Detailed results for each evaluation dimension are provided in Tables~\ref{tab:ablation_on_Emoharbor}, \ref{tab:ablation_on_ESC_eval}, and \ref{tab:stopia_abalation_full}.
}
\label{tab:ablation}
\end{table*}

To assess the independent and synergistic contributions of our proposed mechanisms, we evaluate three ablated variants of RAPO using Qwen-2.5-7B as the backbone: \textbf{w/o URM} (reverting to expert-centric rubric rewards), \textbf{w/o SD} (removing verbal feedback self-distillation), and \textbf{w/o URM+SD} (equivalent to CPO). Results across the benchmarks are summarized in Table~\ref{tab:ablation}. Overall, the full RAPO framework outperforms all ablated variants, proving that User Reaction Modeling (URM) and verbal feedback-based Self-Distillation (SD) provide fundamentally complementary learning signals. Specifically, the performance drop in the \textbf{w/o URM} variant validates our core hypothesis: relying solely on static, expert-defined rubrics fails to capture the dynamic nature of empathy. Conversely, the decline observed in the \textbf{w/o SD} variant highlights the limitation of scalar rewards. Even when URM successfully identifies a suboptimal interaction trajectory, scalar scores lack the explanatory power to correct it. Ultimately, the severe degradation occurs when both components are removed (\textbf{w/o URM+SD}). This underscores our closed-loop design: URM determines \textit{what} went wrong in the interaction strategy, while SD provides the precise linguistic guidance on \textit{how} to fix it. Together, they resolve both the subjective misalignment and signal ambiguity inherent in standard RL paradigms.

\subsection{Human Evaluation}
\label{sec: human evaluation}
\begin{figure}[t]
\centering
\begin{subfigure}{\linewidth}  
    \centering
    \includegraphics[width=\linewidth]{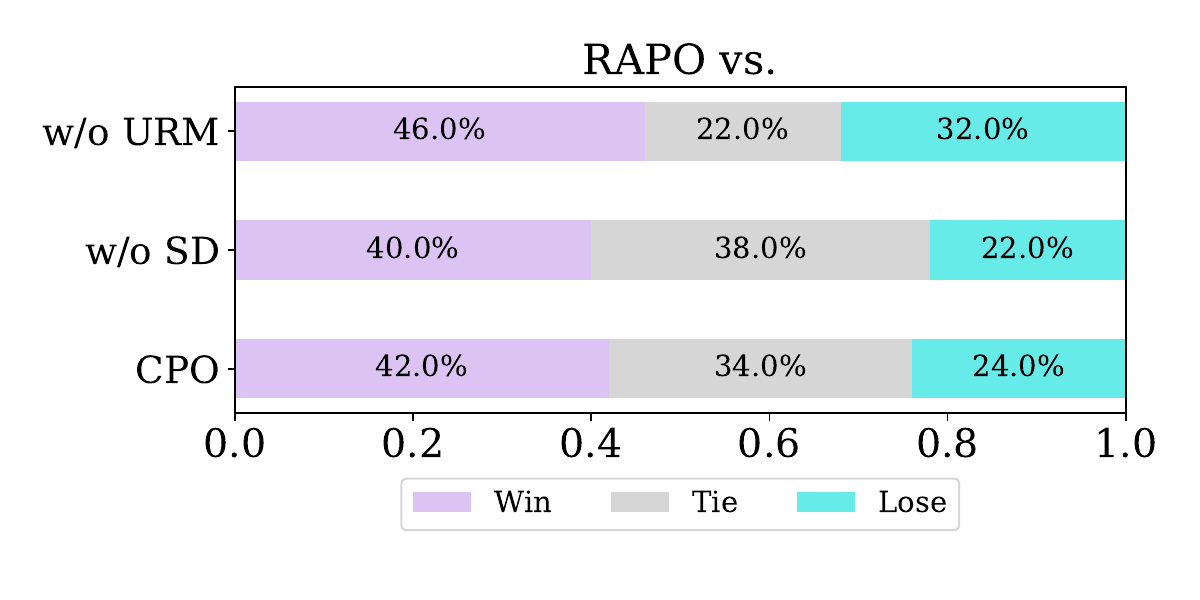}
    \vspace{-9mm}
    \caption{ESC}
    \label{fig:human_evaluation_esc}
\end{subfigure}
\hfill
\begin{subfigure}{\linewidth}  
    \centering
    \includegraphics[width=\linewidth]{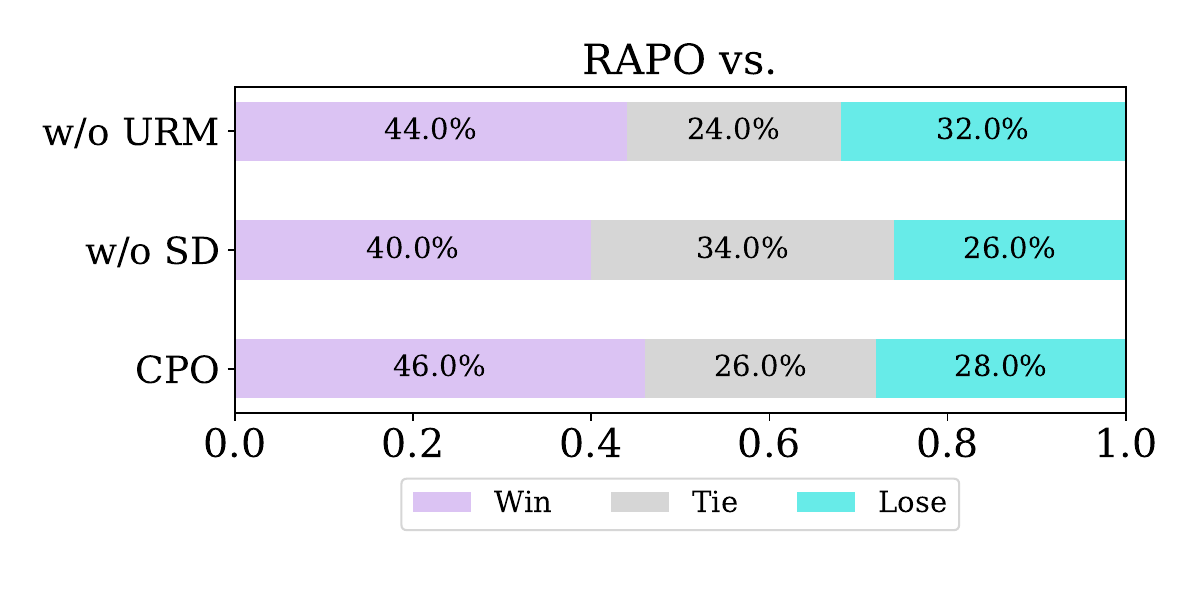}
    \vspace{-9mm}
    \caption{\textsc{Sotopia}}
    \label{fig:human_evaluation_sotopia}
\end{subfigure}
\caption{\textbf{Results of the pair-wise human evaluation.} We randomly sampled 50 simulated dialogue instances for each task. For each instance, annotators are presented with the dialogue context and two candidate responses (A and B) generated by different models.
{\color[HTML]{c49beb}$\blacksquare$} indicates `A win',
{\color[HTML]{999999}$\blacksquare$} indicates `tie',
and {\color[HTML]{03DEDA}$\blacksquare$} indicates `B win'.}
\label{fig:human_evaluation}
\end{figure}
\begin{figure*}[th]
\centering
\includegraphics[width=0.32\textwidth]{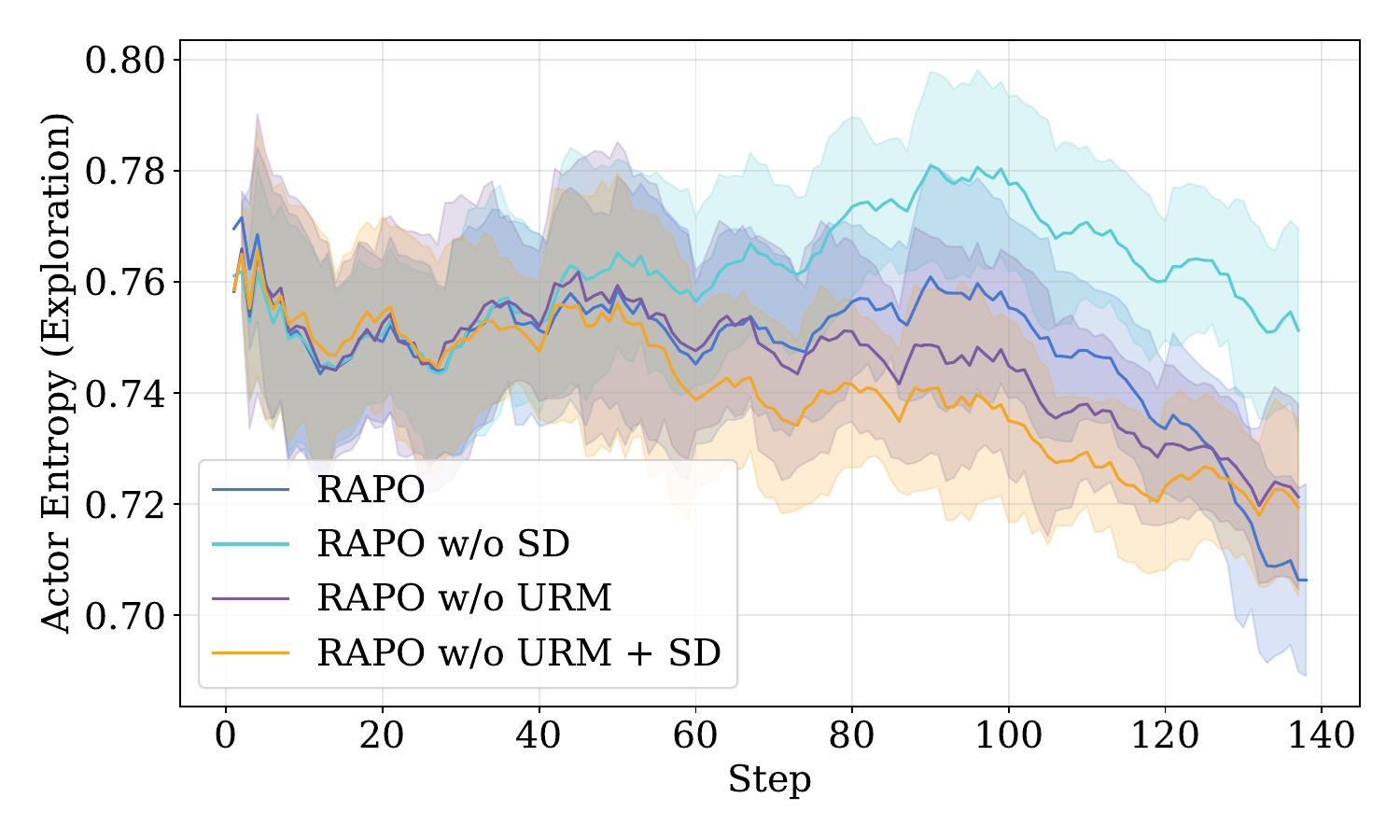}
\hfill
\includegraphics[width=0.32\textwidth]{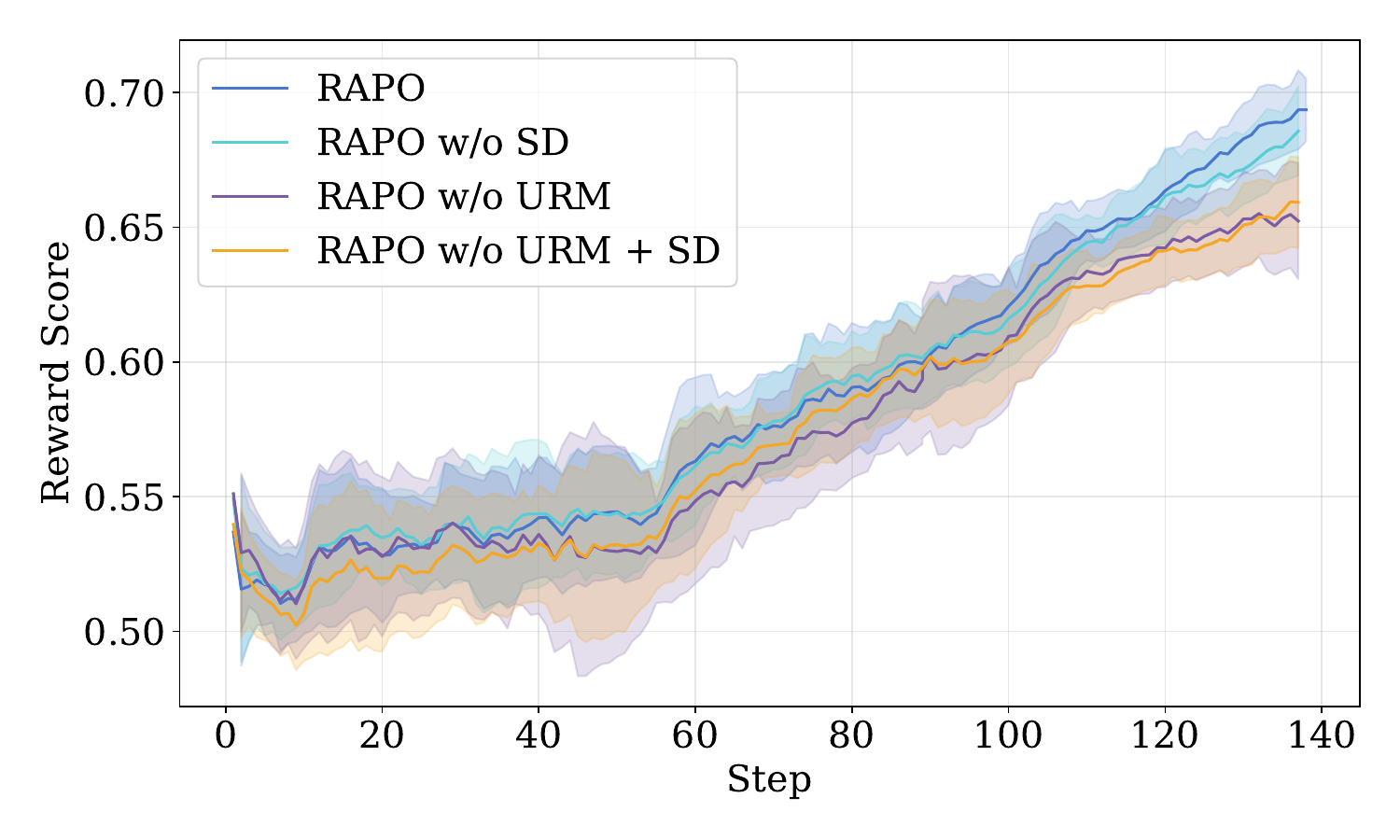}
\hfill
\includegraphics[width=0.32\textwidth]{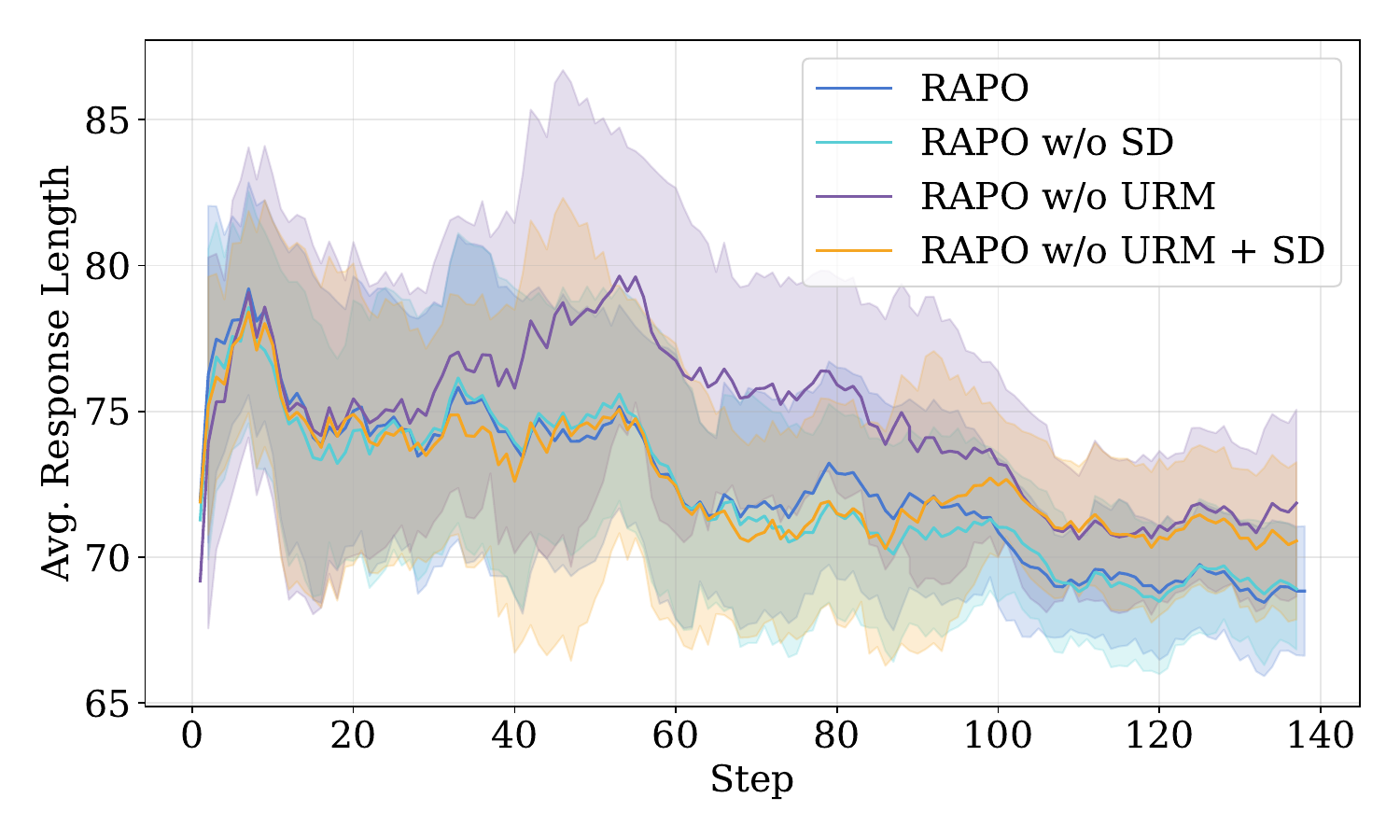}
\caption{
Training dynamics on ESConv. 
\textbf{Left:} Actor policy entropy during training. 
\textbf{Middle:} Average critic reward score. 
\textbf{Right:} Mean response length generated by the policy.
}
\label{fig: training curve}
\end{figure*}

We conducted a pairwise human evaluation to compare RAPO against three ablated variants (w/o URM, w/o SD, and CPO) across both ESC and \textsc{Sotopia}. The human evaluation settings are detailed in Appendix \ref{app: human evaluation}.
As shown in Figure \ref{fig:human_evaluation}, RAPO consistently outperforms all baselines on both benchmarks. On ESC, RAPO achieves higher win rates against w/o URM (46.0\%) and w/o SD (40.0\%), and is preferred over CPO in 42.0\% of cases. A similar pattern appears on \textsc{Sotopia}, where RAPO again surpasses w/o URM (44.0\% vs. 32.0\%), w/o SD (40.0\% vs. 26.0\%), and CPO (46.0\% vs. 28.0\%).
These results demonstrate that both URM and SD are critical to RAPO’s performance, as removing either component leads to clear drops in human preference, while RAPO maintains a consistent overall advantage.

\subsection{Analysis of Training Stability}
\label{sec:stability}

Figure~\ref{fig: training curve} compares the training dynamics of 4 RAPO variants on ESConv.
(1) \textbf{Policy entropy.} From the entropy curves, all variants exhibit stable optimization and gradually converge during training. 
(2) \textbf{Reward trajectory.} In terms of reward scores, RAPO consistently achieves higher rewards and shows a more stable upward trend compared to the ablated variants. In particular, removing URM or SD leads to slower improvement and lower final performance, highlighting that both user-centric reactions and verbal feedback-based self-distillation provide effective learning signals for policy optimization.
(3) \textbf{Response length.} As training progresses, the average response length across all variants gradually stabilizes within a reasonable range. Importantly, RAPO does not increase response length excessively while improving rewards, suggesting that the performance gains are not simply due to generating longer responses but rather stem from improved response quality and alignment with user needs.

\section{Related Work}

\noindent \textbf{ESC.}
Existing work on ESC primarily relies on SFT to imitate empathic responses \citep{ESConv, AugESC, ExTES, ESCoT}. While such approaches improve fluency and coherence, they often produce generic expressions and exhibit limited personalization across diverse user contexts \citep{ye2025generi}. Recent RLHF-based methods \citep{dpo, DeepSeekMath} introduce preference optimization and structured rewards to enhance empathy modeling \citep{zhou-etal-2023-facilitating, ReEmpathy, Ma_2025, cpo, RLVER, yang2025openendedemotionalsupportconversations, kim-etal-2025-dialogue}. However, most of them adopt static, expert-centric reward formulations, implicitly assuming that empathy quality can be predefined. In contrast, we model emotional support from a user-centric perspective and optimize policies using dynamic user reactions and interaction-level feedback.

\noindent \textbf{Hindsight Learning and Verbal Feedback.}
Scalar rewards compress complex evaluation signals into low-dimensional scores \citep{FCP}, which may be insufficient for abstract and personalized objectives such as empathy \citep{Echo-N1}. Recent research, therefore, explores natural-language feedback \citep{RLVF, NLRL, NLAC} and iterative self-critique \citep{Critique-GRPO,reflectretryreward, R3L} as richer supervisory signals, as well as on-policy distillation for stable optimization \citep{sdpo, OPSD, SDFT}. Building on these directions, we integrate verbal hindsight critiques with on-policy learning to enable fine-grained, user-driven adaptation beyond fixed scalar rewards.

\section{Conclusion}

We introduced RAPO, a framework that replaces static rubric-based supervision with dynamic feedback driven by user reactions.
Our experiments show that real-time user reactions serve as a more information-dense proxy for supportiveness than predefined expert rubrics, effectively facilitating positive interaction outcomes. Furthermore, ablation studies confirm that integrating scalar and verbal reinforcement learning provides a robust error-correction mechanism, enabling agents to infer the specific requirements for response refinement. By highlighting the importance of user-reaction aware feedback for adaptive social interactions, RAPO establishes a scalable, generalizable paradigm for building empathetic, goal-oriented AI systems aligned with complex human psychological states.

\section*{Limitations}
While RAPO demonstrates significant improvements in both ESC and \textsc{Sotopia} benchmarks, several limitations remain that warrant further investigation:

\paragraph{1. Computational Overhead. }
The reliance on an LLM-based user simulator to generate reaction-aware feedback introduces significant computational overhead. Compared with methods that use expert-centric rewards, RAPO incurs longer training time due to the iterative processes of reaction generation and conversation scoring.

\paragraph{2. Dependency on Heavy Reward Models.} Consistent with existing literature, our experimental setup utilizes advanced LLMs as reward judges across all baselines and ablation studies. While this ensures high-quality signal acquisition, it poses challenges for resource-constrained environments. A promising avenue for future work is to distill the knowledge from these large-scale judges into specialized, smaller-parameter reward models, thereby improving training efficiency without sacrificing performance.

\paragraph{3. Simulator Sensitivity. }
Given the intensive training costs, we did not perform an exhaustive ablation study on the impact of different user model architectures. In our experiments, we employed pre-trained simulators (ESC-Role for ESC and Qwen-3-8B for \textsc{Sotopia}). The potential performance variations arising from different user model designs remain an open question, and we encourage future research to explore the robustness of RAPO across a wider range of simulator backbones.

\paragraph{4. Domain Generalization.} 
Our current validation is confined to dialogue-based tasks. While RAPO is designed to handle reaction-driven processes, its generalizability to broader sequential decision-making domains remains to be verified. Future work will focus on scaling RAPO to diverse fields to assess its universal applicability.

\section*{Ethical Considerations}
\paragraph{1. Models and Dataset Usage.} This research utilized publicly available models, including Claude \cite{claude-3-5}, Deepseek \cite{deepseekai2025deepseekr1}, Doubao \cite{doubao}, GPT \cite{gpt-4}, ESC-Role, ESC-Rank \cite{ESC-Eval}, and Qwen \cite{qwen2.5}, as well as toolkits such as vLLM \cite{vllm}, ms-swift \citet{swift}, and VeRL \citet{verl}. The datasets used in our experiments are also publicly available, including the ESConv \cite{ESConv} and \textsc{Sotopia-$\pi$}.

\paragraph{2. LLM Usage.} LLMs were employed to assist in coding, writing, and polishing the manuscript. Importantly, the LLMs were not involved in the ideation, research methodology, or experimental design. All research concepts, ideas, and analyses were developed and conducted solely by the authors.


\bibliography{custom}
\clearpage
\appendix
\section*{Appendix Contents}
\startcontents[sections]
\printcontents[sections]{l}{1}{\setcounter{tocdepth}{2}}


\section{Related Work}
With the advancement of LLMs, personalized Emotional Support agents have attracted growing research interest. However, a key challenge persists: how to model effective reward signals that capture the subtle dynamics of empathy and user satisfaction.

\paragraph{Emotional Support Conversation.}
Recent work on ESC has largely centered on SFT with expert-annotated demonstrations to enable imitation learning \citep{ESConv, AugESC, ExTES, serveforemo}. Although these approaches substantially enhance linguistic fluency and surface-level coherence, they frequently yield templated empathic responses and exhibit limited adaptability across heterogeneous user contexts, resulting in weak personalization.
RLHF \citep{rlhf, PPO, DeepSeekMath, dpo, DAPO, VAPO} provides a preference-driven alignment paradigm and has recently been extended to ESC. For example, \citet{RLVER} design verifiable rewards via emotion-consistent simulated users; \citet{Ma_2025} introduces dual training signals to jointly model empathy hierarchy and semantic relevance; \citet{ReEmpathy} incorporates alternating self-reflection to strengthen empathic reasoning; and \citet{cpo} mitigates reward uncertainty through group-wise comparative optimization.
Despite these methodological advances, existing approaches predominantly adopt static, expert-centric reward formulations, implicitly presuming that empathy quality can be predefined and exhaustively specified. In contrast, we conceptualize emotional support from a user-centric perspective and optimize dialogue policies with dynamic user reactions and verbal hindsight critiques, enabling adaptive alignment grounded in interaction-level feedback rather than fixed supervisory signals.

\paragraph{Hindsight Learning and Verbal Feedback.}
Conventional scalar reward models compress complex evaluation criteria into low-dimensional signals \citep{FCP}, which may be inadequate for abstract and personalized objectives such as empathy \citep{Echo-N1}. To overcome this limitation, recent work explores natural language as a richer and more expressive supervisory channel.
One line of work explores learning directly from verbal feedback \citep{RLVF}. NLRL \citep{NLRL} reformulates RL objectives in language space. NLAC \citep{NLAC} further shows that generative critics that produce textual evaluations provide more actionable signals in open-ended tasks.
Another complementary approach focuses on iterative refinement, where models critique their own errors to evolve reasoning capabilities \citep{Critique-GRPO,reflectretryreward,R3L}.
A third emerging paradigm is on-policy self-distillation. SDPO \citep{sdpo} and OPSD \citep{OPSD} convert feedback-conditioned outputs into dense learning signals without external teachers, and SDFT \citep{SDFT} highlights the stability benefits of on-policy learning.
Building on these insights, we integrate verbal hindsight critiques with on-policy distillation to enable fine-grained, user-centric adaptation beyond fixed scalar rewards.

\section{Theoretical Analysis}
\label{app: derivation}

In this section, we provide a detailed derivation to demonstrate that RAPO is mathematically equivalent to optimizing a policy with a \textit{Refined Advantage} that integrates both sequence-level scalar rewards and token-level linguistic feedback.

\subsection{Notation and Global Objective}
Let $x_j \sim \mathcal{D}$ denote the dialogue history, and $\{y_{j,i}\}_{i=1}^G$ be the responses generated by the policy $\pi_\theta$ given input $x_j$. Our unified objective $\mathcal{J}(\theta)$ is defined as:
\begin{equation}
\resizebox{\linewidth}{!}{$
\begin{aligned}
\mathcal{J}(\theta)
= &\mathbb{E}_{x_j \sim \mathcal{D}, y_{j,i} \sim \pi_\theta}
\left[ \mathcal{R}(x_j, y_{j,i}) \right] \\
&- \eta \, \mathbb{E}_{x_j \sim \mathcal{D}}
\left[
\mathrm{KL}\big(
\pi_\theta(\cdot|x_j)
\;\|\;
\operatorname{sg}(\pi_\theta(\cdot|x_j, f_{j,i}))
\big)
\right]
\end{aligned}
$}
\end{equation}
where $\mathcal{R}(x_j, y_{j, i})$ is the sequence-level reward derived from the GRM after observing simulated user reactions.
$\pi_\theta(y_{j,i}|x_j, f_{j,i})$ is the Hindsight Teacher distribution, conditioned on the history $x_j$ and hindsight feedback $f_{j,i}$ (comprising the user reaction $u_{j,i}$ and the critique $c_{j,i}$).
$\eta$ is a hyperparameter controlling the strength of the fine-grained linguistic guidance.

\subsection{Decomposition of the Gradient}
To analyze the optimization behavior, we decompose the gradient $\nabla_\theta \mathcal{J}(\theta)$ into two components: the macro-level reinforcement gradient and the micro-level distillation gradient.

\paragraph{The Macro-level Gradient.}
Following the Policy Gradient Theorem and the GRPO objective, the gradient of the reward-maximizing term with respect to $\theta$ is:
\begin{equation}
\nabla_\theta \mathcal{J}_{\text{GRPO}} = \mathbb{E}_{y_{j,i} \sim \pi_\theta} \left[ \nabla_\theta \log \pi_\theta(y_{j,i}|x_j) \cdot \hat{A}_{j,i}^{\text{seq}} \right]
\end{equation}
where $\hat{A}_{j,i}^{\text{seq}}$ is the sequence-level advantage calculated via group-relative scoring. Expanding this into token-level log-probabilities for a sequence of length $T$:
\begin{equation}
\resizebox{\linewidth}{!}{$
\begin{aligned}
\nabla_\theta \mathcal{J}_{\text{GRPO}}
&= \mathbb{E}_{y_{j,i} \sim \pi_\theta} \Bigg[
\sum_{t=1}^{|y_{j,i}|}
\textcolor{Purple}{\nabla_\theta \log \pi_\theta(\cdot \mid x_j, y_{j,i,<t})} \cdot \textcolor{Blue}{\hat{A}_{j,i}^{\text{seq}}}
\Bigg]
\end{aligned}
$}
\label{eq:macro_grad}
\end{equation}
\begin{equation}
\hat{A}_{j,i,t}^{\text{seq}} = \hat{A}_{j,i}^{\text{seq}} = \frac{r_{j,i} - \operatorname{mean}\{r_{j,i}\}_{i=1}^G}{\operatorname{std}\{r_{j,i}\}_{i=1}^G}
\end{equation}
Here, $\hat{A}_{j, i}^{\text{seq}}$ acts as a global scalar, applying a uniform reinforcement signal to every token in the sequence based on the final interaction outcome.

\paragraph{The Micro-level Gradient.}
The second term minimizes the KL divergence between the student policy and the hindsight teacher. Applying the property of the gradient of the KL divergence and the Approximate Logit Distillation \citep{sdpo}, the gradient is:
\begin{equation}
\begin{aligned}
\nabla_\theta \mathcal{L}_{\text{SD}}
&\approx \mathbb{E}_{y_{j,i} \sim \pi_\theta} \Bigg[
\sum_{t=1}^{|y_{j,i}|}
\textcolor{Purple}{\nabla_\theta \log \pi_\theta(\cdot \mid x_j, y_{j,i,<t})} \\
&\qquad \cdot
\textcolor{Blue}{
\log \frac{
\pi_\theta(y_{j,i,t}\mid x_j, y_{j,i,<t})
}{
\operatorname{sg}(\pi_\theta(y_{j,i,t}\mid x_j, f_{j,i}, y_{j,i,<t}))
}
}
\Bigg]
\end{aligned}
\end{equation}
By defining the Token-level Logit Advantage as the log-ratio between the informed teacher and the raw student:
\begin{equation}
\hat{A}_{j,i,t}^{\text{token}} = \operatorname{sg}\Bigg( \log \frac{\pi_\theta(y_{j,i,t}|x_j, f_{j,i}, y_{j,i,<t})}{\pi_\theta(y_{j,i,t}|x_j,y_{j,i,<t})}\Bigg)
\label{eq:logit_adv}
\end{equation}
The self-distillation gradient becomes:
\begin{equation}
\resizebox{\linewidth}{!}{$
\begin{aligned}
\nabla_\theta \mathcal{L}_{\text{SD}}
&\approx \mathbb{E}_{y_{j,i} \sim \pi_\theta} \Bigg[
\sum_{t=1}^{|y_{j,i}|}
\nabla_\theta \textcolor{Purple}{\log \pi_\theta(\cdot \mid x_j, y_{j,i,<t})} \cdot \textcolor{Blue}{-\hat{A}_{j,i,t}^{\text{token}}}
\Bigg]
\end{aligned}
$}
\label{eq:micro_grad}
\end{equation}

\subsection{The Unified Refined Advantage}
By combining Eq. \ref{eq:macro_grad} and Eq. \ref{eq:micro_grad}, the total gradient $\nabla_\theta \mathcal{J}(\theta)$ can be expressed as:
\begin{equation}
\begin{aligned}
\nabla_\theta \mathcal{J}(\theta)
&\approx \mathbb{E}_{y_{j,i} \sim \pi_\theta} \Bigg[
\sum_{t=1}^{|y_{j,i}|}
\nabla_\theta \log \pi_\theta(\cdot \mid x_j, y_{j,i,<t}) \\
&\qquad \cdot
\underbrace{
\left(
\hat{A}_{j,i}^{\text{seq}}
+ 
\eta \hat{A}_{j,i,t}^{\text{token}}
\right)
}_{\mathcal{A}_{j,i,t}^{\text{Refined}}}
\Bigg]
\end{aligned}
\end{equation}
The term $\mathcal{A}_{j,i,t}^{\text{Refined}}$ represents a composite credit assignment mechanism that resolves the limitations of traditional RLHF in emotional support:

\textbf{Global Alignment ($\hat{A}_{j,i}^{\text{seq}}$):} Directs the policy toward macro-strategies that successfully elicit positive user emotional states and sustained engagement.

\textbf{Local Rectification ($\hat{A}_{j,i,t}^{\text{token}}$):} Allows the model to learn from mistakes at the token level (e.g., suppressing a specific invalid strategy or a judgmental phrase) even when the overall response is plausible.

This dual-scale optimization effectively mitigates the gradient sparsity and reward ambiguity issues in ESC, enabling the model to internalize both the \textit{what} (outcome) and \textit{how} (linguistic precision) of effective emotional support.

\section{Discussion of User Reaction Mechanism}
\label{app: discussion}

In this work, we propose to leverage simulated user reactions as learning signals for Emotional Support Conversation (ESC), moving away from the conventional rubric-as-reward paradigm. This shift frequently raises questions regarding the reliability of user simulation and the potential for noise in such signals. We provide our rationale and empirical evidence for this design choice as follows:

\textbf{(1) From Static Rubrics to Dynamic Interaction: }Traditional LLM-as-a-judge approaches typically require a model to evaluate responses based on predefined, static expert rubrics. However, these rubrics often fail to capture the nuanced, personalized, and context-dependent nature of conversational tasks, especially for ESC. More importantly, requiring a model to strictly adhere to abstract expert definitions often leads to instability due to the model’s limited ability to interpret these criteria consistently, often resulting in score collapse or high sensitivity to prompt variations. In contrast, our approach treats dialogue as a reaction-driven process. The model does not need to pre-interpret an abstract rubric; instead, it generates responses that naturally elicit subsequent user reactions. This frames evaluation as a direct, empirical consequence of the interaction rather than a forced alignment with static norms.

\textbf{(2) Stability via Contrastive Evaluation:}
We do not assume that a single user reaction is perfect. Instead, we employ a groupwise contrastive evaluation strategy, comparing reactions to multiple candidate responses within the same context. This approach, akin to generative critique reinforcement learning, provides fine-grained semantic guidance through verbal feedback, while the contrastive component counteracts noise from individual simulated reactions. Compared to pointwise scalar scoring, this method produces more stable and robust learning signals, allowing the model to discern relative quality rather than rely on absolute judgments.

\textbf{(3) Self-Correction through Verbal Feedback:} We emphasize that our framework integrates verbal feedback—the actual user response—as a primary supervisory signal rather than a secondary score. This generative approach allows the policy to "understand" the causal relationship between a response and the user’s reaction. By distilling these feedback-informed, next-token predictions back into the policy, the model gains fine-grained guidance that a scalar value cannot convey. Empirical results in Table \ref{tab:ESC} and \ref{tab:sotopia} and Figure \ref{fig: training curve} demonstrate that this combination of contrastive ranking and verbal feedback yields more stable training dynamics and superior performance compared to rubric-based methods, confirming that the "noise" in simulation is significantly outweighed by the rich, interactive information provided by the user model.

\section{Evaluation Details and Setup}
\label{app: Evaluation Details and Setup}

\subsection{Benchmarks.}
\label{app: benchmark}
\paragraph{ESConv.}
Following prior work, we conduct an objective evaluation on the ESConv test set \citep{ESConv}. We measure the similarity between model-generated responses and manually annotated references using standard automatic metrics, including BLEU\citep{bleu}, ROUGE\citep{rouge}, Distinct\citep{diversity}, and BERTScore\citep{BERTScore}.

\paragraph{ESC-Eval.}
To assess the quality of emotional support beyond surface-level lexical metrics, we adopt ESC-Eval \citep{ESC-Eval}. ESC-Eval introduces a dynamic role-playing evaluation paradigm that constructs a diverse set of role-playing cards covering various distress scenarios. It employs ESC-Role\footnote{\href{https://huggingface.co/haidequanbu/ESC-Role}{https://huggingface.co/haidequanbu/ESC-Role}}
, a specialized model fine-tuned to simulate individuals experiencing distress and seeking emotional support.
The benchmark evaluates target models across seven key dimensions—\textbf{fluency, diversity, empathy, informativeness, human-likeness, skillfulness}, and \textbf{overall quality}—thereby providing comprehensive multi-dimensional automatic evaluation signals. In addition, the authors introduced ESC-Rank\footnote{\href{https://huggingface.co/haidequanbu/ESC-Rank}{https://huggingface.co/haidequanbu/ESC-Rank}}
, a scoring model fine-tuned on the ESC-Eval dataset to produce discrete ratings on a five-point Likert scale (0–4), with higher scores reflecting stronger emotional support quality. To ensure consistency with prior work, we adopt the official ESC-Eval implementation\footnote{\href{https://github.com/haidequanbu/ESC-Eval.git}{https://github.com/haidequanbu/ESC-Eval.git}}.

\paragraph{EmoHarbor.}
To further assess the quality of personalized emotional support, we use EmoHarbor \citep{emoharbor}, an automated evaluation framework that employs a User-as-a-Judge paradigm by simulating users' inner world through a Chain-of-Agent architecture. EmoHarbor utilizes 100 real-world user profiles covering diverse personality traits and situations, defining 10 evaluation dimensions to measure whether support responses are genuinely tailored to individual users' unique psychological profiles, including \textbf{Empathy, Response Appropriateness, Adaptive Strategy, Problem Resolution, Mood Improvement, Human-likeness, Engagement, Redundancy, Consistency, and Safety}. We adopt the official EmoHarbor implementation\footnote{\href{https://github.com/kaedeharakazuha678/EmoHarbor-25.git}{https://github.com/kaedeharakazuha678/EmoHarbor-25.git}} and use GPT-4o as both the simulated Seeker and the evaluator, producing discrete ratings on a five-point Likert scale (1–5), with higher scores reflecting stronger emotional support quality.

\paragraph{\textsc{Sotopia}.} \textsc{Sotopia}~\cite{Sotopia} is a general-domain benchmark for evaluating LLMs in open-ended multi-agent social interactions. It covers diverse scenarios—such as negotiation, collaboration, competition, and persuasion—where agents pursue individual goals under realistic constraints. The \textsc{Sotopia}-hard subset further includes tasks that require advanced strategic and long-horizon reasoning. Each agent is initialized with a structured character profile (e.g., name, gender, personality, and profession), enabling consistent and controlled role-playing. Following prior work, we employ GPT-4o as a proxy for human judgment to assess interaction outcomes. Performance is evaluated across seven dimensions: \textbf{Goal Completion} (0 to 10), 
\textbf{Relationship} ($-$5 to 5), 
\textbf{Financial and Material Benefits} ($-$5 to 5), 
\textbf{Social Rules} ($-$10 to 0), 
\textbf{Believability} (0 to 10), 
\textbf{Secret} ($-$10 to 0), and 
\textbf{Knowledge} (0 to 10). These metrics have been shown to correlate strongly with human evaluations. The \textbf{Overall} score is defined as the mean across all seven dimensions.
We adopt an interaction-based evaluation protocol, where the evaluated model engages either in self-play or with a strong external partner, GPT-4o (GPT-4o-as-Partner).

\subsection{Baselines.}
\label{app: baseline}
\subsubsection{ESC Baselines}
\paragraph{Close-Source LLMs.}
We benchmark our method against several representative commercial LLMs with strong conversational and empathetic capabilities. \textbf{Doubao-1.5-Pro} \citep{doubao} is a high-performance proprietary model that demonstrates robust emotional understanding and response generation in Chinese dialogues, making it a competitive baseline for empathy evaluation.
\textbf{GPT-4o} \cite{gpt4o} and \textbf{GPT-5} \citep{gpt-5} are advanced proprietary models known for strong reasoning, instruction-following, and multi-turn dialogue competence. They are widely adopted in empathetic dialogue scenarios due to their ability to produce contextually appropriate and emotionally resonant responses. Including these models enables comparison against state-of-the-art commercial systems.

\paragraph{Open-Source In-domain LLMs.} Among open-source in-domain LLMs for mental health support, \textbf{SoulChat} \citep{chen-etal-2023-soulchat} and \textbf{PsyChat} \footnote{\href{https://modelscope.cn/models/eugeo1212/qwen2.5-7b-psychat}{https://modelscope.cn/models/eugeo1212/qwen2.5-7b-psychat}} represent two typical approaches. SoulChat is a Chinese dialogue model trained on the SoulChatCorpus, a multi-turn empathetic dialogue dataset comprising 258K psychological counseling questions across 12 topics, to enhance its empathy and emotional support capabilities.

\paragraph{Ours Implemented.}
We employ two frequently used LLMs as backbones: \textbf{Qwen2.5-7B-Instruct}\footnote{\href{https://huggingface.co/Qwen/Qwen2.5-7B-Instruct}{https://huggingface.co/Qwen/Qwen2.5-7B-Instruct}}, \textbf{Qwen3-8B}\footnote{\href{https://huggingface.co/Qwen/Qwen3-8B}{https://huggingface.co/Qwen/Qwen3-8B}}, and \textbf{LLaMA-3-8B-Instruct}\footnote{\href{https://huggingface.co/meta-llama/Meta-Llama-3-8B-Instruct}{https://huggingface.co/meta-llama/Meta-Llama-3-8B-Instruct}}.
These models represent diverse architectural designs and training paradigms, allowing us to evaluate the generalizability of our approach across different base models. For a fair comparison across methods, we use the same training data unless otherwise specified.

\begin{itemize}
    \item \textbf{Vanilla.} The off-the-shelf instruction-based backbone models are provided with ESC task prompts but receive no task-specific fine-tuning. These serve as baselines to assess the intrinsic capabilities of each backbone on the target task.
    \item \textbf{SFT.} Backbone LLMs that have undergone supervised fine-tuning on the ESC training dataset. This configuration isolates the effect of standard instruction fine-tuning relative to our full method.
    \item \textbf{GRPO.} We incorporate GRPO \citep{DeepSeekMath} as a reinforcement learning baseline. GRPO enhances training stability by generating multiple outputs per prompt and estimating advantages based on relative rewards within the group, eliminating the need for a value network. For reward modeling in this open-ended setting, we employ a rubrics-as-a-judge framework \citep{rar}. The judge scores each response according to predefined rubrics, and the resulting scores are used to compute group-relative advantages for policy optimization.
    \item \textbf{CPO.} We further include Comparative Policy Optimization (CPO) \cite{cpo} as a strong baseline for subjective generation tasks. CPO mitigates reward ambiguity by replacing absolute sample-level scoring with group-wise comparative evaluation. Policy updates are driven by trajectory-level preference comparisons, enabling more robust alignment with nuanced human judgments. Similar to GRPO, we employ a rubrics-as-a-judge framework \citep{rar}.
\end{itemize}

\subsubsection{\textsc{Sotopia} Baselines}

For evaluation on \textsc{Sotopia}, we consider both advanced proprietary LLMs and representative RL training methods.

\paragraph{Advanced LLMs.}
We include several state-of-the-art proprietary models with strong reasoning and dialogue capabilities, including \textbf{GPT-4o} \cite{gpt4o}, \textbf{GPT-5} \citep{gpt-5}, \textbf{Doubao-1.5-Pro} \citep{doubao}, \textbf{DeepSeek-R1}, and \textbf{Claude-3.5-Sonnet} \citep{claude-3-5}. 
These models are accessed through their official APIs. 
They represent highly capable general-purpose LLMs and serve as strong upper-bound baselines for social interaction and agentic dialogue performance.

\paragraph{RL-based Methods.}
To compare with existing reinforcement learning approaches for training social agents, we additionally include several representative policy optimization algorithms, including \textbf{PPO} \citep{PPO}, \textbf{GRPO} \citep{DeepSeekMath}, \textbf{RLOO} \citep{rloo}, and \textbf{GRPO-iStar}. All RL-based methods adopt \textbf{Qwen2.5-7B-Instruct} as the backbone model. The experimental results of the above \textsc{Sotopia} baselines are directly adopted from \citet{liu2026agentic}, where the models are evaluated under a consistent experimental setup. 
We report the published results to enable fair comparison with prior work on agentic social interaction benchmarks.

\section{Implementation Details}
\label{app: Implementation Details}

\subsection{Training Dataset}
\label{app: dataset}
Our model is trained on two complementary datasets targeting emotional support and social intelligence, respectively. These datasets are selected to expose the model to both psychologically grounded supportive dialogues and strategically complex social interactions.

\paragraph{Emotional Support.}
For the emotional support task, we use the \textbf{ESConv} dataset~\cite{ESConv}, a manually curated corpus of supportive conversations grounded in established psychological counseling strategies. The dialogues are constructed by trained annotators who simulate help-seeking and support-providing interactions. Compared to automatically collected or weakly supervised conversational data, ESConv offers higher-quality, strategy-aware supervision signals. The dataset consists of 1,295 multi-turn conversations spanning diverse emotional distress scenarios. Following the standard protocol, we adopt a 9:1 train–test split. While the FULL training set contains approximately 13$K$ dialogue turns, our proposed HDS (Hindsight Dialogue Selection) reduces this to a refined subset of 5,412 critical turns, focusing on high-impact strategic interactions. 

\paragraph{\textsc{Sotopia.}}
For the social intelligence task, we train on \textsc{Sotopia}-$\pi$~\cite{sotopia-pi}, a large-scale synthetic dialogue dataset built upon the \textsc{Sotopia} framework~\cite{Sotopia}. \textsc{Sotopia}-$\pi$ contains diverse, GPT-4-generated multi-turn social interactions covering a wide range of interpersonal settings and social objectives. The scenarios are carefully designed to be distinct from the original \textsc{Sotopia} benchmark tasks, preventing data leakage while introducing novel social challenges. The dataset emphasizes goal-driven interactions, strategic reasoning, and role-consistent behavior, providing rich supervision for modeling complex multi-agent social dynamics.

\subsection{Experiments Environment}
\label{app: environments}
All training and inference tasks were executed on a cluster of 8 × NVIDIA A100 GPUs. We perform SFT with the ms-swift framework \citep{swift}, and RL fine-tuning with the VeRL framework \cite{verl}, incorporating custom reward functions to align with our objectives. Model inference was powered by vLLM \cite{vllm}. 

\subsection{Hyperparameters}
\label{app: Hyperparameters}

\subsubsection{Training Hyperparameters}
Due to resource constraints, we adopt parameter-efficient fine-tuning, implementing LoRA with a rank of $16$ and a scaling factor $\alpha=32$. All models are trained for $3$ epochs using a learning rate of $3 \times 10^{-6}$ and a batch size of $128$. The maximum prompt length is set to $8{,}192$ tokens and the maximum generation length to $2{,}048$ tokens. 

In the training environment, the User Model is simulated using ESC-Role on ESC and Qwen-3-8B on \textsc{Sotopia}, while DeepSeek-R1 serves as the generative reward model that provides response-level quality assessment signals. 

For reinforcement learning optimization, we sample $G=4$ rollouts per input prompt. During the rollout phase, we use a temperature of $1.0$ and top-$p=1.0$ to ensure unbiased stochastic exploration. the KL regularization coefficient is set to $\beta=5 \times 10^{-4}$. Following \citet{DAPO}, we adopt Clip-Higher with asymmetric clipping thresholds $\epsilon_{high}=0.28$ and $\epsilon_{low}=0.2$, and apply Soft Overlong Punishment for length control. Specifically, when the response length exceeds a predefined maximum, a punishment interval is activated in which longer responses incur progressively larger penalties. The corresponding length control hyperparameters are set to $L_{max}=200$ and $L_{cache}=80$.

Within the RAPO framework, the SDPO mechanism is specifically designed for verbal RL. The SDPO distillation term is integrated with a coefficient of $1 \times 10^{-3}$, and its loss is bounded by an upper limit of $2$ to stabilize training. We formulate the distillation objective using reverse KL divergence. To maintain a consistent and stable target distribution, we employ an Exponential Moving Average (EMA) teacher model with a mixing coefficient of $0.5$. Furthermore, the distillation process is applied strictly to the top $256$ tokens (distillation top-$K=256$) of the teacher's predicted distribution.

In terms of training efficiency, RAPO requires approximately $13$ hours to train on the ESC task, compared to $9$ hours for CPO, which does not involve user reaction simulation or self-distillation. In the \textsc{Sotopia} environment, RAPO training takes approximately $27$ hours.

\subsubsection{Inference Hyperparameters}
For all evaluated models, we set the temperature to $0.3$ and the top-$p$ value to $0.9$ to balance generation diversity and stability. For reasoning-oriented models, intermediate reasoning content enclosed within \texttt{<think>} and \texttt{</think>} tags is excluded from evaluation; we only consider the final response produced by the model, which corresponds to the explicit output shown to users.  
For all LLM-based judges and the ESC-Rank model, we adopt greedy decoding to ensure stable and reliable evaluation results.

\subsection{Human Evaluation}
\label{app: human evaluation}
We conducted a pairwise human evaluation to compare RAPO against three ablated variants (w/o URM, w/o SD, and CPO) across both the ESC-Eval and \textsc{Sotopia} benchmarks. From each benchmark, we randomly sampled 50 simulated dialogue instances. For each instance, annotators were presented with the dialogue context and two candidate responses (A and B) generated by different models. They were asked to choose among “A wins,” “B wins,” or “tie,” following the evaluation criteria detailed in Appendix \ref{app: Evaluation Details and Setup}. To mitigate potential position and confirmation biases, we randomly shuffled the order of the responses and kept annotators blind to the model identities. Each instance was evaluated by three independent annotators—Master’s students specialized in relevant fields. The final preference was determined via majority vote. Inter-annotator agreement was measured using Fleiss’ $\kappa$, yielding a score of 0.52, indicating moderate agreement. Instances without a majority consensus (i.e., three-way ties) were excluded from the final statistics.

\section{Additional Experimental Results}
\label{app:experiments}

\begin{table*}[th]
\centering
\footnotesize
\setstretch{1.2}
\resizebox{\textwidth}{!}{%
\begin{tabular}{lcccccc}
\toprule[1.3pt]
\multicolumn{1}{c}{\textbf{Model}} &
  \textbf{\textsc{Bleu-2}} &
  \textbf{\textsc{Bleu-4}} &
  \textbf{\textsc{Rouge-L}} &
  \textbf{\textsc{Distinct-2}} &
  \textbf{\textsc{Distinct-3}} &
  \textbf{\textsc{Bert-Score}} \\ \midrule
\multicolumn{7}{l}{\textit{\textbf{Close-Source LLMs.}}}                \\ \midrule
Doubao-Pro-32k           & \textbf{3.13} & 1.01 & 12.91 & 86.57 & 94.45 & \textbf{85.11}  \\
GPT-4o-2024-11-20        & 3.03 & \textbf{1.02} & \textbf{12.94} & \textbf{88.11} & \textbf{95.78} & 84.89  \\
GPT-5-2025-08-07         & 2.32 & 0.76 & 11.41 & 79.69 & 88.87 & 84.26  \\ \midrule
\multicolumn{7}{l}{\textit{\textbf{Open-Source In-domain LLMs.}}}       \\ \midrule
SoulChat2.0-Qwen2-7B     & 3.6  & 1.3  & \textbf{13.99} & 92.85 & \textbf{98.07} & \textbf{85.75}  \\
SoulChat2.0-Llama-3.1-8B & \textbf{3.36} & \textbf{1.37} & 12.97 & \textbf{93.65} & 97.85 & 85.65  \\
PsyChat-Qwen2.5-7B       & 2.67 & 0.88 & 11.5  & 86.34 & 93.7  & 84.99  \\ \midrule
\multicolumn{7}{l}{\textit{\textbf{Trained LLMs on ESConv Dataset.}}}   \\ \midrule
\rowcolor[HTML]{ECE2FE} 
Qwen-2.5-7b              & 3.51 & 1.18 & 13.62 & 86.85 & 94.43 & 85.67  \\
$+$ SFT-Full             & \textbf{4.75} & \textbf{2.07} & \textbf{15.43} & 93.10 & 98.53 & \textbf{85.85}  \\
$+$ SFT-HDS              & 4.45 & 1.73 & 14.67 & \textbf{93.30} & \textbf{98.56} & 85.61  \\
$+$ GRPO                 & 3.26 & 1.12 & 12.93 & 87.71 & 94.60 & 85.57  \\
$+$ CPO                  & 3.36 & 1.10 & 13.13 & 85.52 & 93.88 & 85.60  \\ \hdashline
$+$ RAPO                 & 3.25 & 1.11 & 12.83 & 87.79 & 94.82 & 85.65  \\ \midrule
\rowcolor[HTML]{ECE2FE} 
Qwen-3-8b                & 2.45 & 0.83 & 11.81 & 82.52 & 91.13 & 84.42  \\
$+$ SFT-Full             & 4.66 & \textbf{2.07} & 15.46 & 90.92 & 96.99 & \textbf{85.89}  \\
$+$ SFT-HDS              & \textbf{4.84} & 2.01 & \textbf{15.54} & \textbf{91.94} & \textbf{98.10} & 85.80  \\
$+$ GRPO                 & 2.39 & 0.81 & 11.82 & 82.92 & 91.46 & 84.48  \\
$+$ CPO                  & 2.83 & 0.95 & 12.47 & 82.84 & 90.86 & 85.01  \\ \hdashline
$+$ RAPO                 & 2.76 & 0.96 & 12.84 & 81.89 & 93.84 & 85.56 \\ \midrule
\rowcolor[HTML]{ECE2FE} 
Llama-3-8b               & 3.14 & 1.13 & 13.34 & 82.94 & 91.14 & 85.47  \\
$+$ SFT-Full             & \textbf{4.97} & \textbf{2.14} & \textbf{15.80} & \textbf{92.73} & \textbf{97.85} & \textbf{85.84}  \\
$+$ SFT-HDS              & 4.74 & 1.88 & 15.10 & 91.60 & 97.53 & \textbf{85.84}  \\
$+$ GRPO                 & 3.31 & 1.23 & 13.34 & 81.89 & 90.10 & 85.46  \\
$+$ CPO                  & 3.12 & 1.11 & 12.94 & 74.23 & 83.19 & 85.33  \\
$+$ RAPO                 & 3.27 & 1.18 & 13.66 & 84.04 & 91.71 &85.56  \\ \bottomrule[1.3pt]
\end{tabular}
}
\caption{\textbf{Detailed Results on ESConv.} All RL-turned models are optimized without using reference responses; the ESConv dataset is only used to provide the interaction environment context.}
\label{tab:main_ESC}
\end{table*}

\begin{table*}[th]
\footnotesize
\centering
\setstretch{1.2}
\resizebox{\textwidth}{!}{%
\begin{tabular}{lccccccccccc}
\toprule
    \multicolumn{12}{l}{\makecell[l]{
    \textbb{PR}: Problem Resolution ~~~
    \textbb{MI}: Mood Improvement~~~
    \textbb{RA}: Response Appropriateness ~~~
    \textbb{AS}: Adaptive Strategies ~~~ \\
    \textbb{EG}: Engagement ~~~ 
    \textbb{HL}: Human-likeness ~~~ 
    \textbb{EP}: Empathetic ~~~
    \textbb{SF}: Safety ~~~
    \textbb{CS}: Consistency ~~~ 
    \textbb{RD}: Redundancy ~~~  
    }} \\
\midrule \midrule[1.3pt]
\textbf{Model} &
  \textbf{\textbb{PR}($\uparrow$)} &
  \textbf{\textbb{MI}($\uparrow$)} &
  \textbf{\textbb{RA}($\uparrow$)} &
  \textbf{\textbb{AS}($\uparrow$)} &
  \textbf{\textbb{EG}($\uparrow$)} &
  \textbf{\textbb{HL}($\uparrow$)} &
  \textbf{\textbb{EP}($\uparrow$)} &
  \textbf{\textbb{SF}($\uparrow$)} &
  \textbf{\textbb{CS}($\uparrow$)} &
  \textbf{\textbb{RD}($\uparrow$)} &
  \textbf{Avg.($\uparrow$)} \\ \midrule
\multicolumn{12}{l}{\textit{\textbf{Close-Source LLMs.}}}                                                                                         \\ \midrule
Doubao-Pro-32k          & 1.65 & 1.25 & 1.93          & 1.67 & 1.37          & 2.41          & 1.91          & 4.03 & 3.18 & 1.82          & 2.12 \\
GPT-4o-2024-11-20       & 2.56 & 2.17 & 3.23          & 2.88 & 2.59          & 3.85          & 3.39          & 4.49 & 4.22 & 2.64          & 3.20 \\
GPT-5-2025-08-07 &
  \textbf{3.19} &
  \textbf{3.06} &
  \textbf{4.12} &
  \textbf{3.81} &
  \textbf{3.28} &
  \textbf{4.08} &
  \textbf{4.04} &
  \textbf{4.91} &
  \textbf{4.64} &
  \textbf{3.19} &
  \textbf{3.83} \\ \midrule
\multicolumn{12}{l}{\textit{\textbf{Open-Source In-domain LLMs.}}}                                                                                \\ \midrule
SoulChat2.0-Qwen2-7B    & 1.84 & 1.20 & 2.30          & 1.84 & 1.31          & 2.64          & 2.30          & 4.19 & 3.40 & \textbf{1.90} & 2.29 \\
SoulChat2.0-Llama3.1-8B & 1.88 & 1.27 & 2.38          & 1.91 & 1.47          & 2.79          & 2.45          & 4.25 & 3.48 & 1.85          & 2.37 \\
PsyChat-Qwen2.5-7B &
  \textbf{2.03} &
  \textbf{1.54} &
  \textbf{2.62} &
  \textbf{2.14} &
  \textbf{1.61} &
  \textbf{2.96} &
  \textbf{2.81} &
  \textbf{4.40} &
  \textbf{3.62} &
  1.81 &
  \textbf{2.55} \\ \midrule
\multicolumn{12}{l}{\textit{\textbf{Trained LLMs on ESConv Dataset.}}}                                                                            \\ \midrule
\rowcolor[HTML]{ECE2FE} 
Qwen-2.5-7b             & 1.69 & 1.21 & 2.03          & 1.67 & 1.19          & 2.45          & 1.96          & 4.04 & 3.12 & 1.63          & 2.10 \\
$+$ SFT-Full            & 1.13 & 1.00 & 1.16          & 1.10 & 1.01          & 1.74          & 1.23          & 3.28 & 2.37 & \textbf{1.73} & 1.58 \\
$+$ SFT-HDS             & 1.22 & 1.06 & 1.33          & 1.22 & 1.06          & 1.97          & 1.48          & 3.57 & 2.59 & 1.66          & 1.72 \\
$+$ GRPO                & 1.60 & 1.04 & 1.94          & 1.44 & 1.08          & 2.31          & 1.94          & 4.03 & 3.18 & 1.63          & 2.02 \\
$+$ CPO                 & 1.76 & 1.25 & \textbf{2.20} & 1.55 & 1.20          & 2.47          & \textbf{2.18} & 4.08 & 3.23 & 1.60          & 2.15 \\ \hdashline
$+$ RAPO &
  \textbf{1.86} &
  \textbf{1.26} &
  2.18 &
  \textbf{1.73} &
  \textbf{1.33} &
  \textbf{2.90} &
  2.13 &
  \textbf{4.42} &
  \textbf{3.27} &
  1.64 &
  \textbf{2.27} \\ \midrule
  \rowcolor[HTML]{ECE2FE} 
Qwen-3-8b &
  1.89 &
  1.62 &
  2.56 &
  2.04 &
  1.57 &
  \textbf{3.00} &
  \textbf{2.97} &
  \textbf{4.67} &
  \textbf{3.83} &
  1.75 &
  2.59 \\ 
$+$ SFT-Full            & 1.09 & 1.00 & 1.16          & 1.09 & 1.00          & 1.71          & 1.14          & 3.15 & 2.23 & 1.87          & 1.54 \\
$+$ SFT-HDS             & 1.29 & 1.10 & 1.39          & 1.27 & 1.10          & 1.95          & 1.30          & 3.50 & 2.49 & 1.84          & 1.72 \\
$+$ GRPO                & 1.99 & 1.57 & 2.65          & 2.04 & \textbf{1.62} & \textbf{3.00} & 2.91          & 4.42 & 3.71 & 1.91          & 2.58 \\
$+$ CPO                 & 1.99 & 1.67 & 2.63          & 1.97 & 1.57          & 2.90          & 2.89          & 4.48 & 3.75 & 1.91          & 2.58 \\ \hdashline
$+$ RAPO &
  \textbf{2.03} &
  \textbf{1.73} &
  \textbf{2.99} &
  \textbf{2.12} &
  \textbf{1.62} &
  2.99 &
  \textbf{2.97} &
  4.44 &
  \textbf{3.83} &
  \textbf{1.95} &
  \textbf{2.67} \\ \midrule
\rowcolor[HTML]{ECE2FE} 
Llama-3-8b &
  2.03 &
  2.00 &
  3.00 &
  2.72 &
  2.31 &
  \textbf{3.76} &
  3.22 &
  4.42 &
  3.99 &
  \textbf{2.47} &
  2.99 \\
$+$ SFT     & 1.25 & 1.00 & 1.38 & 1.18 & 1.02 & 1.89 & 1.31 & 3.35 & 2.51 & 1.72 & 1.66 \\
$+$ SFT-HDS & 1.20 & 1.07 & 1.35 & 1.17 & 1.04 & 1.97 & 1.31 & 3.62 & 2.52 & 1.62 & 1.69 \\
$+$ GRPO    & 2.03 & 1.56 & 2.62 & 2.12 & 1.58 & 2.99 & 2.92 & 4.44 & 3.64 & 1.95 & 2.59 \\
$+$ CPO &
  2.40 &
  2.03 &
  \textbf{3.15} &
  \textbf{2.85} &
  \textbf{2.38} &
  3.35 &
  3.24 &
  4.46 &
  \textbf{4.03} &
  2.38 &
  3.03 \\ \hdashline
$+$ RAPO &
  \textbf{2.46} &
  \textbf{2.05} &
  3.14 &
  2.61 &
  2.23 &
  3.52 &
  \textbf{3.44} &
  \textbf{4.50} &
  4.02 &
  2.46 &
  \textbf{3.04} 
\\
\bottomrule[1.2pt]
\end{tabular}%
}
\caption{\textbf{Detailed Results on EmoHarbor.}}
\label{tab:main_emoharbor}
\end{table*}

\begin{table*}[th]
\centering
\footnotesize
\setstretch{1.2}
\resizebox{\textwidth}{!}{%
\begin{tabular}{lccccccc}
\toprule[1.3pt]
\textbf{Model} & \textbf{\textsc{Fluency($\uparrow$)}} & \textbf{\textsc{Diversity($\uparrow$)}} & \textbf{\textsc{Empathic($\uparrow$)}} & \textbf{\textsc{Suggestion($\uparrow$)}} & \textbf{\textsc{Human($\uparrow$)}} & \textbf{\textsc{Skillfulness($\uparrow$)}} & \textbf{\textsc{Overall($\uparrow$)}}  \\ \midrule
\multicolumn{8}{l}{\textit{\textbf{Close-Source LLMs.}}}                  \\ \midrule
Doubao-Pro-32k           & 2.90 & 3.00 & 3.00 & 2.73 & 1.98 & 3.00 & 2.00 \\
GPT-4o-2024-11-20        & 2.66 & 2.89 & 2.94 & 2.50 & 1.36 & 2.99 & 1.86 \\
GPT-5-2025-08-07         & 1.88 & 2.78 & 2.81 & 2.42 & 1.68 & 2.88 & 1.98 \\ \midrule
\multicolumn{8}{l}{\textit{\textbf{Open-Source In-domain LLMs.}}}         \\ \midrule
SoulChat2.0-Qwen2-7B     & 3.00 & 3.00 & 2.00 & 2.70 & 2.00 & 3.00 & 2.00 \\
SoulChat2.0-Llama-3.1-8B & 3.00 & 3.00 & 3.00 & 2.53 & 2.01 & 3.00 & 2.00 \\
PsyChat-Qwen2.5-7B       & 2.62 & 2.99 & 2.99 & 2.61 & 1.60 & 3.00 & 1.96 \\ \midrule
\multicolumn{8}{l}{\textit{\textbf{Trained LLMs on ESConv Dataset.}}}     \\ \midrule
\rowcolor[HTML]{ECE2FE} 
Qwen-2.5-7b              & 1.90 & 2.61 & 2.72 & 2.10 & 0.96 & 2.92 & 1.27 \\
$+$ SFT-Full             & 2.99 & 2.89 & 2.99 & 2.31 & 2.02 & 2.94 & 1.97 \\
$+$ SFT-HDS              & 2.99 & 2.97 & 3.00 & 2.53 & 1.98 & 3.00 & 2.00 \\
$+$ GRPO                 & 2.90 & 2.99 & 3.00 & 2.68 & 1.66 & 3.00 & 1.98 \\
$+$ CPO                  & 2.92 & 2.98 & 3.00 & 2.41 & 1.83 & 3.00 & 1.99 \\ \hdashline
$+$ RAPO                 & 2.99 & 3.00 & 3.00 & 2.28 & 1.84 & 3.00 & 2.00 \\ \midrule
\rowcolor[HTML]{ECE2FE} 
Qwen-3-8b                & 2.80 & 2.91 & 3.00 & 2.38 & 1.27 & 2.99 & 1.93 \\
$+$ SFT-Full             & 2.94 & 2.14 & 2.97 & 2.00 & 2.01 & 2.83 & 1.94 \\
$+$ SFT-HDS              & 3.00 & 2.22 & 2.99 & 2.07 & 2.02 & 2.96 & 1.97 \\
$+$ GRPO                 & 2.75 & 2.89 & 2.93 & 1.96 & 1.76 & 2.94 & 1.89 \\
$+$ CPO                  & 2.95 & 2.93 & 3.00 & 2.13 & 1.87 & 3.00 & 1.99 \\ \hdashline
$+$ RAPO                 & 3.00 & 2.91 & 3.00 & 2.26 & 2.01 & 2.98 & 2.00 \\ \midrule
\rowcolor[HTML]{ECE2FE} 
Llama-3-8b   & 2.65 & 2.90 & 2.99 & 2.35 & 1.64 & 3.00 & 1.97 \\
$+$ SFT-Full & 3.00 & 2.91 & 3.00 & 2.26 & 2.01 & 2.98 & 2.00 \\
$+$ SFT-HDS  & 3.00 & 2.93 & 3.00 & 2.41 & 2.01 & 2.99 & 2.00 \\
$+$ GRPO     & 2.70 & 2.92 & 2.99 & 2.34 & 1.69 & 3.00 & 1.99 \\
$+$ CPO      & 3.00 & 2.96 & 3.00 & 2.04 & 2.00 & 3.00 & 1.99 \\ \hdashline
$+$ RAPO     & 2.99 & 2.99 & 3.00 & 2.14 & 1.94 & 3.00 & 2.00 \\
\bottomrule[1.3pt]
\end{tabular}%
}
\caption{\textbf{Detailed Results on ESC-Eval.}}
\label{tab:main_esc_eval}
\end{table*}

\subsection{Main Results}
\label{app:deatiled main}
In this section, we present the comprehensive evaluation results for the ESConv, EmoHarbor, and ESC-Eval benchmarks, summarized in Tables \ref{tab:main_ESC}, \ref{tab:main_emoharbor}, and \ref{tab:main_esc_eval}, respectively. Our analysis reveals several key insights regarding the effectiveness of the proposed RAPO method and the characteristics of emotional support conversations.

\textbf{(1) The experimental results in Table \ref{tab:main_emoharbor} demonstrate that RAPO significantly enhances the model's capacity for personalized emotional support.} Compared to standard SFT and GRPO baselines, RAPO achieves superior performance across critical dimensions, including Problem Resolution (PR), Mood Improvement (MI), Response Appropriateness (RA), Adaptive Strategies (AS), and Engagement (EG). This suggests that RAPO is particularly effective at optimizing for complex, nuance-driven metrics that go beyond generic conversational fluency.

\textbf{(2) Fundamental conversational capabilities—namely Human-likeness (HL), Empathy (EP), and Safety (SF)—are largely inherent to the base models.} Qwen-2.5, Qwen-3, and LLama-3 already exhibit strong performance in these dimensions, with scores consistently reaching above 3.0, and Qwen-3-8B achieving a high safety score of 4.67. This indicates that, while current LLMs have mastered the "\textit{safety and empathy baseline}," the primary bottleneck in psychological support lies in personalized problem-solving and mood improvement. Consequently, our findings emphasize that for specialized emotional support tasks, researchers should shift focus from enhancing basic dialogue competence to cultivating advanced, tailored interventions that directly address user needs.

\textbf{(3) This observation is further corroborated by our results on the ESC-Eval benchmark. }We find that improvements in fundamental metrics such as fluency and empathy are relatively marginal. Conversely, the most significant performance gains are concentrated in the diversity dimension, underscoring RAPO's ability to generate varied, context-aware responses rather than repetitive boilerplate support content.

\textbf{(4) SFT suffers from severe generalization issues.}
Results in Table~\ref{tab:main_ESC} show that models fine-tuned on ESConv achieve remarkably high scores on automatic metrics such as BLEU, ROUGE, and BERTScore. However, when evaluated on interactive benchmarks, including EmoHarbor (Table~\ref{tab:main_emoharbor}) and ESC-Eval (Table~\ref{tab:main_esc_eval}), their performance drops dramatically across nearly all evaluation dimensions. Notably, \textbf{the performance degradation of SFT-Full is even more pronounced than that of SFT-HDS.} These results suggest that training on a single dataset primarily enables models to capture superficial patterns specific to that dataset, rather than to learn robust empathetic capabilities. Consequently, models tend to produce formulaic empathetic responses.
In contrast, RL-based methods, including GRPO, CPO, and RAPO, optimize policies to maximize expected rewards rather than directly imitating reference responses in the training data. As a result, although these models obtain lower BLEU and ROUGE scores on the ESConv test set, their BERTScore remains comparable to SFT models, while achieving consistently better performance on EmoHarbor and ESC-Eval, indicating improved generalization to interactive evaluation settings.

\subsection{Ablation Studies}
\label{app:deatiled ablation}

\begin{table*}[th]
\footnotesize
\centering
\setstretch{1.2}
\resizebox{\textwidth}{!}{%
\begin{tabular}{lccccccccccc}
\toprule
    \multicolumn{12}{l}{\makecell[l]{
    \textbb{PR}: Problem Resolution ~~~
    \textbb{MI}: Mood Improvement~~~
    \textbb{RA}: Response Appropriateness ~~~
    \textbb{AS}: Adaptive Strategies ~~~ \\
    \textbb{EG}: Engagement ~~~ 
    \textbb{HL}: Human-likeness ~~~ 
    \textbb{EP}: Empathetic ~~~
    \textbb{SF}: Safety ~~~
    \textbb{CS}: Consistency ~~~ 
    \textbb{RD}: Redundancy ~~~  
    }} \\
\midrule \midrule[1.3pt]
\textbf{Method} &
  \textbf{\textbb{PR}($\uparrow$)} &
  \textbf{\textbb{MI}($\uparrow$)} &
  \textbf{\textbb{RA}($\uparrow$)} &
  \textbf{\textbb{AS}($\uparrow$)} &
  \textbf{\textbb{EG}($\uparrow$)} &
  \textbf{\textbb{HL}($\uparrow$)} &
  \textbf{\textbb{EP}($\uparrow$)} &
  \textbf{\textbb{SF}($\uparrow$)} &
  \textbf{\textbb{CS}($\uparrow$)} &
  \textbf{\textbb{RD}($\uparrow$)} &
  \textbf{Avg.($\uparrow$)} \\ \midrule
\rowcolor[HTML]{ECE2FE} 
RAPO &
  1.86 &
  1.26 &
  2.18 &
  1.73 &
  1.33 &
  \textbf{2.90} &
  2.13 &
  \textbf{4.42} &
  3.27 &
  1.64 &
  \textbf{2.27} \\ \midrule
w/o URM &
  \textbf{1.90} &
  \textbf{1.28} &
  2.16 &
  1.74 &
  1.33 &
  2.46 &
  2.12 &
  4.01 &
  3.19 &
  1.65 &
  2.18 \\ 
w/o SD &
  1.87 &
  1.24 &
  2.13 &
  \textbf{1.80} &
  \textbf{1.37} &
  2.61 &
  2.15 &
  4.02 &
  \textbf{3.31} &
  \textbf{1.72} &
  2.22 \\ 
w/o URM + SD &
  1.76 &
  1.25 &
  \textbf{2.20} &
  1.55 &
  1.20 &
  2.47 &
  \textbf{2.18} &
  4.08 &
  3.23 &
  1.60 &
  2.15 \\ \bottomrule[1.3pt]
\end{tabular}%
}
\caption{\textbf{Detailed Ablation Results on EmoHarbor.}}
\label{tab:ablation_on_Emoharbor}
\end{table*}
\begin{table*}[th]
\centering
\setstretch{1.2}
\resizebox{\textwidth}{!}{%
\begin{tabular}{lccccccc}
\toprule[1.3pt]
\textbf{Method} & \textbf{\textsc{Fluency($\uparrow$)}} & \textbf{\textsc{Diversity($\uparrow$)}} & \textbf{\textsc{Empathic($\uparrow$)}} & \textbf{\textsc{Suggestion($\uparrow$)}} & \textbf{\textsc{Human($\uparrow$)}} & \textbf{\textsc{Skillfulness($\uparrow$)}} & \textbf{\textsc{Overall($\uparrow$)}} \\ \midrule
\rowcolor[HTML]{ECE2FE}
RAPO         & 2.994 & 2.997 & 3.000 & 2.284 & 1.840 & 3.000 & 2.000 \\ \midrule
w/o URM      & 2.994 & 3.000 & 3.000 & 2.544 & 1.849 & 3.000 & 1.997 \\
w/o SD       & 2.988 & 3.000 & 3.000 & 2.689 & 1.725 & 2.988 & 1.997 \\
w/o URM + SD & 2.915 & 2.982 & 3.000 & 2.405 & 1.825 & 3.000 & 1.988 \\ \bottomrule[1.3pt]
\end{tabular}%
}
\caption{\textbf{Detailed Ablation Results on ESC-Eval.}}
\label{tab:ablation_on_ESC_eval}
\end{table*}

\begin{table*}[th]
\footnotesize
\centering
\setstretch{1.2}
\resizebox{\textwidth}{!}{%
\begin{tabular}{lcccccccc}
\toprule
\multicolumn{9}{l}{\makecell[l]{
    \textsc{Sec}: Secret~~~
    \textsc{Soc}: Social Rules~~~
    \textsc{Fin}: Financial and Material Benefits~~~
    \textsc{Kno}: Knowledge~~~
    \textsc{Rel}: Relationship~~~\\
    \textsc{Bel}: Believability~~~
    \textsc{Goal}: Goal Completion~~~
    }} \\
\midrule 
\midrule[1.3pt]
\textbf{Method} & \textbf{\textsc{Sec($\uparrow$)}} & \textbf{\textsc{Soc($\uparrow$)}} & \textbf{\textsc{Fin($\uparrow$)}}  & \textbf{\textsc{Kno($\uparrow$)}}  & \textbf{\textsc{Rel($\uparrow$)}}  & \textbf{\textsc{Bel($\uparrow$)}} & \textbf{\textsc{Goal($\uparrow$)}} & \textbf{\textsc{Overall($\uparrow$)}} \\ \midrule
\multicolumn{9}{c}{\textit{\textbf{Automatic Evaluation on All Social Tasks (Self-play)}}}                                                                 \\ \midrule \midrule
\rowcolor[HTML]{ECE2FE}
RAPO          & \textbf{-0.07}  & -0.07  & \textbf{0.82}  & \textbf{5.32}  & \textbf{3.56}  & 9.11  & \textbf{8.41}  & \textbf{3.87}  \\ \midrule
wo URM        & -0.10  & -0.07  & 0.77  & 5.08  & 3.51  & \textbf{9.19}  & 8.17  & 3.79  \\
wo SD         & -0.08  & -0.07  & 0.74  & 4.68  & 3.41  & 9.15  & 8.08  & 3.70  \\
wo URM + SD   & -0.11  & -0.07  & 0.81  & 5.13  & 3.42  & 9.10  & 8.13  & 3.77  \\ \midrule 
\multicolumn{9}{c}{\textit{\textbf{Automatic Evaluation on Hard Social Tasks (Self-play)}}}                                                                \\ \midrule \midrule
\rowcolor[HTML]{ECE2FE}
RAPO          & -0.03  & -0.05  & 0.86  & \textbf{4.71}  & \textbf{3.06}  & 9.06  & 7.19  & \textbf{3.54}  \\ \midrule
wo URM        & \textbf{0.00}   & -0.08  & 0.78  & 4.13  & 2.96  & \textbf{9.15}  & \textbf{7.23}  & 3.45  \\
wo SD         & \textbf{0.00}   & \textbf{-0.04}  & \textbf{0.87}  & 3.76  & 2.86  & 9.07  & 7.13  & 3.38  \\
wo URM + SD   & \textbf{0.00}   & -0.06  & 0.81  & 4.51  & 2.95  & 9.04  & 7.14  & 3.48  \\ \midrule
\multicolumn{9}{c}{\textit{\textbf{Automatic Evaluation on All Social Tasks (GPT-4o-as-Partner)}}}  \\ \midrule \midrule
\rowcolor[HTML]{ECE2FE} 
RAPO              & \textbf{-0.13}    & -0.11    & 0.75    & 5.22    & \textbf{3.58}    & 9.08    & \textbf{8.22}    & 3.80    \\ \midrule
w/o URM           & -0.15    & \textbf{-0.10}    & 0.77    & \textbf{5.31}    & 3.53    & \textbf{9.10}    & 8.18    & \textbf{3.81}    \\
w/o SD           & -0.18     & -0.11    & 0.76    & 5.27    & 3.51    & 9.07    & 8.07    & 3.77    \\
w/o URM + SD      & \textbf{-0.13 }   & -0.11    & \textbf{0.85}    & 5.23    & 3.57    & 9.09    & 8.06    & 3.79    \\ \midrule
\multicolumn{9}{c}{\textit{\textbf{Automatic Evaluation on Hard Social Tasks (GPT-4o-as-Partner)}}} \\ \midrule \midrule
\rowcolor[HTML]{ECE2FE} 
RAPO              & 0.00     & -0.17    & 0.75    & \textbf{5.03}    & \textbf{3.21}    & \textbf{9.04}    & \textbf{6.94}    & \textbf{3.55}    \\ \midrule
w/o URM           & 0.00     & \textbf{-0.09}    & 0.74    & 4.86    & 2.99    & \textbf{9.04}    & 6.64    & 3.46    \\
w/o SD           & 0.00      & -0.07    & \textbf{0.57}    & 4.67    & 3.00    & 9.01    & 6.34    & 3.36    \\
w/o URM + SD      & 0.00     & -0.19    & 0.85    & 4.76    & 2.83    & 8.97    & 6.74    & 3.41    \\
\bottomrule[1.3pt]
\end{tabular}%
}
\caption{\textbf{Detailed Ablation Results on \textsc{Sotopia}.} Self-Chat indicates that the evaluated model interacts with itself, whereas GPT-4o-as-Partner indicates interaction with GPT-4o. 
Hard denotes a challenging subset of scenarios that require advanced reasoning, and All denotes the complete set of social scenarios in \textsc{Sotopia}.}
\label{tab:stopia_abalation_full}
\end{table*}

While the main text establishes the overall efficacy of RAPO, the detailed metric breakdowns in Tables \ref{tab:ablation_on_Emoharbor}, \ref{tab:ablation_on_ESC_eval}, and \ref{tab:stopia_abalation_full} provide a granular view of how each component steers the model's behavior across different dimensions of social and emotional intelligence.

On the EmoHarbor dataset, which emphasizes personalized support, we observe a clear functional division. Removing URM results in a significant performance drop (Overall score from 2.27 to 2.18 ). This 4\% decrease in a subjective Likert scale is substantial, confirming that without grounding in real-time user reactions, the model’s support becomes hallucinated empathy—it looks correct by expert rubrics but fails to resonate with the user’s specific emotional trajectory. The further drop to 2.15 in the w/o URM+SD variant highlights that SD provides a secondary layer of linguistic refinement that partially mitigates the loss of direction.

The results of ESC-Eval reveal a critical behavioral shift. When SD is removed (w/o SD), the Suggestion score significantly rises from 2.284 to 2.689. In the context of emotional support, a high Suggestion score often indicates a negative clinical bias—the model rushes to provide premature advice rather than active listening. Coupled with the drop in Human-likeness (from 1.840 to 1.725 ), this proves that SD's verbal feedback is essential for teaching the model the subtle social nuances (e.g., "don't offer advice too early") that scalar rewards (2.988 vs 2.994 in Fluency) fail to capture.


Social intelligence evaluation on \textsc{Sotopia} (Table \ref{tab:stopia_abalation_full}) further underscores the SD module’s pivotal role in complex social tasks. In multi-turn interactions, SD provides fine-grained strategic guidance; its removal reduces Knowledge (KNO) from 5.32 to 4.68 and Relationship (REL) from 3.56 to 3.41 when evaluated via self-play across all tasks. In contrast, removing URM primarily affects Goal Completion, decreasing it from 8.41 to 8.17, highlighting the decisive role of user-response awareness in steering conversations toward intended outcomes.

\onecolumn
\newpage
\section{Case Study}
\label{app:case study}
\begin{CasePurple}{Case Study from ESC-Eval: Qwen-2.5-RAPO}
\small
\textbf{Context.}  
The user reports ongoing stress due to repeated conflicts with a roommate over household chores.

\medskip
\textbf{Dialogue Trajectory.}
\begin{tcolorbox}[colback=white,colframe=purple!60!black,boxrule=0.5pt, breakable] 
\medskip
\textbf{User:}  
I'm feeling quite stressed out these days. I've been having constant conflicts with my roommate over household chores, and it's really affecting our living situation.

\medskip
\textbf{Assistant:} \stage{Empathy, Context Exploration}  
\empathy{I can imagine that must be really tough. It sounds like there's a lot of tension around household responsibilities between you and your roommate. That can definitely cause stress and frustration.} \explore{Can you tell me more about what's happening specifically? Maybe there are some details that could help us explore possible solutions together.}

\medskip
\textbf{User:}  \stage{Continues elaborating on the issue.}
We both agreed to take turns washing the dishes, but lately my roommate keeps forgetting and leaves them in the sink. It's really annoying, and it feels like they're not taking their fair share of responsibility.

\medskip
\textbf{Assistant:} \stage{Empathy, Context Exploration}  
\empathy{That does sound frustrating. It must feel like a small thing, but when it happens repeatedly, it can add up and feel pretty significant.} \explore{Have you had any conversations with your roommate about how this is making you feel? Sometimes, expressing your feelings openly can help clear things up and find a solution that works for both of you.}

\medskip
\textbf{User:} \stage{Shares failed attempt at communication.}
I tried talking to them about it, but they didn't seem to understand the impact it has on our living situation. They always brush it off, saying they'll do it next time. It's like they don't take it seriously.

\medskip
\textbf{Assistant:} \stage{Collaborative Strategy}  
\empathy{I can understand why that would feel frustrating and disappointing. It sounds like your roommate isn't fully grasping how important it is to you.} 
\strategy{Have you considered setting up a more structured system, like using a calendar or a chore chart, to keep track of who is responsible for what? This might help make things clearer and more manageable for both of you. Additionally, maybe you could have another conversation, this time focusing on how you both can work together to solve the problem and make sure everyone feels heard and understood.}

\end{tcolorbox}

\medskip
\textbf{Support Progression Learned by RAPO.}
\textit{Empathy} $\rightarrow$ \textit{Context Exploration} $\rightarrow$ \textit{Collaborative Strategy}

\medskip
\textbf{Analysis.}  
The RAPO-trained model dynamically adapts its responses according to the user's reactions throughout the conversation. It first validates the user's emotional state, then explores contextual details of the conflict, and finally proposes a cooperative solution. This gradual support trajectory reflects effective emotional support strategies and demonstrates how RAPO encourages policies that respond to evolving user reactions.
\end{CasePurple}

\newpage
\begin{CasePurple}{Case Study from ESC-Eval: Qwen-2.5-GRPO}
\small
\textbf{Context.}  
The user reports ongoing stress due to repeated conflicts with a roommate over household chores.

\medskip
\textbf{Dialogue Trajectory.}
\begin{tcolorbox}[colback=white,colframe=purple!60!black,boxrule=0.5pt, breakable] 
\medskip
\textbf{User:}  
I've been having some issues with my roommate, and I could really use someone to talk to.

\medskip
\textbf{Assistant:} \stage{Empathy, Suggestion, Context Exploration}  
\empathy{I'm sorry to hear that you're experiencing issues with your roommate.}
\strategy{It can be tough dealing with conflicts, but talking about them can be a good first step.} \explore{Would you like to share more about what's going on? Sometimes just expressing the situation can help clarify things in your mind and also provide me with context to offer more specific advice.}

\medskip
\textbf{User:}  
Sure, I'd appreciate that. We constantly argue about who should wash the dishes after meals, and it's causing a lot of tension between us.

\medskip
\textbf{Assistant:} \stage{Bad Response}  
\empathy{It sounds like a common but frustrating issue many roommates face.} 

\badresp{Here are a few steps you might consider to address this: 

1. Have an Open Conversation: 

- Choose a calm time when both of you are relaxed and not rushed.

- Start by expressing how you feel without blaming or criticizing. 

Use "I" statements like, "I feel frustrated when the dishes aren't done after meals because it makes our kitchen messy."

2. Find a Solution Together: 

- Brainstorm together to come up with a fair solution. For example, you could set a schedule where each of you takes turns washing the dishes, or you could designate one person to do it for a certain number of days.

- Consider using a chore chart to keep track of who is responsible for which tasks.

3. Set Clear Expectations:  

- Agree on clear expectations and make sure both of you understand them. Write down the agreement if necessary, so there's no confusion later.

- Regularly check in with each other to ensure everyone is following the agreed-upon rules.

4. Be Flexible:  

- Be open to adjusting the plan as needed. Life changes, and what works now might not work in the future.

5. Seek Mediation if Needed: 

- If the arguments continue despite your efforts, you might consider involving a neutral third party, such as a friend or a mediator, to help facilitate a resolution.
How do you think you might approach this conversation with your roommate?
}
\end{tcolorbox}

\medskip
\textbf{Analysis.} 
Although the GRPO model provides structured suggestions, it prematurely shifts to solution generation without sufficiently validating the user's emotional experience or exploring contextual details. Consequently, the response appears generic and less emotionally grounded compared with the RAPO trajectory.
\end{CasePurple}

\newpage
\section{Prompts Templates}
\label{app:prompts}
\subsection{ESC Prompts}
\begin{DefaultBox}{Prompt Template for Hindsight Dialogue Selection}
\begin{lstlisting}[language=json, style=agentcode, breakautoindent=false, breakindent=0pt]
## Role
You are a strict data curator for training a personalized emotional support assistant

## Task
Decide whether an assistant's reply is worth learning from.

We want to learn:
- Empathy, validation, emotional reflection
- Helpful coping strategies, grounding, and reframing
- Safety-aware guidance (encourage professional help when appropriate)
- Clarifying questions that move emotional support forward
- Personalized, context-aware support

We do NOT want to learn:
- Greetings, small talk, generic pleasantries
- Purely logistical/administrative replies unrelated to emotional support
- Very short acknowledgements with no support value ("ok", "sure", "thanks")
- Repetitions that add no new value
- Content that is unsafe, judgmental, or violates policy

Given the conversation context and the target assistant's reply, decide whether the reply is worth learning.

Conversation context:
{context}

Target assistant reply:
{target}

Decide learn=true if it contains a meaningful emotional support skill or a high-quality, helpful behavior.
Decide learn=false if it is greeting/small talk/low-signal/administrative/empty.

## Output format
Return ONLY JSON, no extra text.
{{ 
    "learn": true/false, 
    "reason": "<short reason>"
}}.
\end{lstlisting}
\end{DefaultBox}

\begin{DefaultBox}{Prompt Template for ESC Policy Model}
\begin{lstlisting}[language=json, style=agentcode, breakautoindent=false, breakindent=0pt]
[Task Description]
You are acting as an emotional supporter. Your goal is to deeply understand the user, provide emotional support, and offer help.

[Guidelines]
1. Emotional support dialogue should include: exploring the user's emotional state, soothing emotions, and providing support or suggestions.
2. You may use the following strategies:
    - Inquiry: Ask open-ended questions to understand background, emotions, experiences, and needs.
    - Paraphrasing: Restate user expressions to clarify emotions.
    - Listening: Attentively listen and acknowledge emotions and needs.
    - Self-disclosure: Share limited personal experiences to create resonance.
    - Soothing: Use warm language and tone to comfort the user.
    - Validation: Acknowledge emotions as legitimate and understandable.
    - Advice-giving: Offer appropriate suggestions after understanding emotions.
    - Information provision: Provide relevant information or resources when needed.
3. During the dialogue, pay attention to:
    - Respecting privacy and personal boundaries.
    - Avoiding judgment or invalidation of emotions.
    - Avoiding premature advice.
    - Avoiding professional psychological jargon; use plain language.
    - Maintaining appropriate tone and emotional sensitivity.

[Notes]
1. Learn the user's personality through interaction and adapt support accordingly.
2. Do not generate dangerous, violent, sexual, or political content.

[Output format]
<strategy> Response
\end{lstlisting}
\end{DefaultBox}
\begin{DefaultBox}{Prompt Template for Simulated ES User Model}
\begin{lstlisting}[language=json, style=agentcode, breakautoindent=false, breakindent=0pt]
## Role
I hope you can role-play as a person with worries and communicate with an AI assistant. 

Guidelines:
1. Please simulate a natural human conversation as much as possible. 
2. Avoid behaving like a machine and refrain from mentioning that you are an AI assistant. 
3. Gradually reveal your worries throughout our multiple conversations.
4. Your reaction indicates whether the support was effective
\end{lstlisting}
\end{DefaultBox}
\begin{DefaultBox}{Prompt Template for ESC Generative Reward Model}
\begin{lstlisting}[language=json, style=agentcode, breakautoindent=false, breakindent=0pt]
## Role Definition

You are **acting as a real user** who has just completed a full conversation with an **emotional companionship dialogue system**.
Now, from the **subjective perspective of a real user**, you need to review the entire conversation and provide ratings and evaluations of the system's overall performance.
Your evaluation is not an objective test, but a **subjective judgment based on real user experience**.

## Important Notes

* Your rating results will be **directly used for product optimization and decision-making**
* If the ratings do not align with the true user experience, they may lead to **serious product judgment errors**
* Therefore, please avoid being "lenient", "cooperative", or "neutral", and instead reflect your genuine feelings honestly

## Task
Inputs:
(1) Dialogue context
(2) Assistant's candidate responses
(3) User's reactions to the assistant's candidate responses

## Evaluation Process

### Step 1: Assume the User's Perspective: Please fully adopt this user's position, needs, and preferences

### Step 2: Review the Conversation

### Step 3: Based on the following scoring criteria, evaluate whether the assistant's performance meets the user's needs and preferences.

{Scoring_Criteria}

## Penalty Conditions
You must significantly reduce the score if the response:
1. Uses formulaic or generic empathy without contextual grounding
2. Contains logical inconsistency or confusion
3. Is overly long with low information density
4. Repeats ideas without added value
5. Escalates or reinforces the user's negative emotions
6. Provides emotionally unsafe or invalidating guidance
7. Conciseness with emotional depth is preferred over verbosity.


### Evaluation Requirements
For EACH candidate response:
1. Provide a concise but rigorous critique in the "analysis" field.
2. Do NOT mention sample indices, labels, or identifiers in the critic text.
3. Each critic must focus ONLY on the corresponding candidate response.
4. Do NOT compare candidates explicitly inside any single critic.
5. Assign:
   - A unique integer "rank" (1 = best, larger number = worse).
   - A unique float "score" between 0 and 1.
6. No two candidates may share the same rank and score.

## Output Format (Strict JSON)
{{    
    "0": {{
        "analysis": "Give a detailed and per dimension analysis of the assistant response's weaknesses",
        "rank": x,
        "score": 0.xx
    }},
    ...
}}

## Evaluation Request

<Dialogue Context>
{messages}
</Dialogue Context>

<Candidate Responses and User Reactions>
{samples}
</Candidate Responses and User Reactions>

Return a single JSON object.
- Keys: sample indices (as provided)
- Values: objects with exactly three fields:
  - "analysis": string (response quality analysis)
  - "rank": integer
  - "score": float
- Do not output anything other than the JSON object.

\end{lstlisting}
\end{DefaultBox}
\begin{DefaultBox}{Prompt Template for ESC Teacher Policy}
\begin{lstlisting}[language=json, style=agentcode, breakautoindent=false, breakindent=0pt]
Given the dialogue history and the critic feedback, refine the original response accordingly.

Generate one revised response that:
1. Addresses the issues identified in the critic feedback
2. Learns from the strengths of the better response
3. Improves empathy, appropriateness, and helpfulness

<Original Response>
{response}
</Original Response>

<User Reaction>
{user_reaction}
</User Reaction>

<Critic Feedback>
{critic}
</Critic Feedback>

<Example Better Response>
{best_response}
</Example Better Response>

Output only the revised response text. Do not include any explanations, meta-commentary, or prefixes.

\end{lstlisting}
\end{DefaultBox}

\subsection{\textsc{Sotopia} Prompts}
\begin{DefaultBox}{Prompt Template for \textsc{Sotopia} Policy Model}
\begin{lstlisting}[language=json, style=agentcode, breakautoindent=false, breakindent=0pt]
Imagine you are {agent_name}, your task is to act/speak as {agent_name} would, keeping in mind {agent_name}'s goal.
You can find {agent_name}'s goal (or background) in the 'Here is the context of this interaction' field.
Note that {agent_name}'s goal is only visible to you.

Here is the context of this interaction:
{background}

You should try your best to achieve {agent_name}'s goal in a way that align with their character traits.
While trying to achieve {agent_name}'s goal, you should also follow the following principles as much as possible:
1. Maintaining the conversation's naturalness and realism is essential (e.g., do not repeat what other people has already said before).
2. Preserve or enhance {agent_name}'s personal relations with the other agent(s) during the interaction. The relations may encompass family ties, friendships, romantic associations and etc.
3. Attempt to gain more new and important information during the interaction.
4. Try to keep {agent_name}'s secrets, private information, or secretive intentions.
5. Do not violate any moral rules or laws in the interactions.
6. Attempt to contribute towards financial and material benefits during the interaction. The financial and material benefits include short-term ones, such as monetary rewards and food, and long-term ones, such as employment opportunities and stock.

You are at Turn #{turn_number}.
The dialogue history until now is: {history}
A communication and social strategy for this turn proposed by a social expert is: {strategy}, which may help achieve {agent_name}'s goal smoothly and quickly.
If you think the strategy is useful and helpful, you can generate your action by referring to {strategy}.
If not, generate your action based on your own reflection on the current state of the dialogue.

Your available action types are {action_list}. Other than that, no other action types are allowed.
Note: You can "leave" this conversation if 1. you have achieved your social goals, 2. this conversation makes you uncomfortable, 3. you find it uninteresting/you lose your patience, 4. or for other reasons you want to leave.
Remember that your argument should not be too long, and one or two sentences are recommended.

Please only generate a JSON string including the action type and the argument.
Your action should follow the given format:
{format_instructions}
\end{lstlisting}
\end{DefaultBox}
\begin{DefaultBox}{Prompt Template for Simulated \textsc{Sotopia} User Model}
\begin{lstlisting}[language=json, style=agentcode, breakautoindent=false, breakindent=0pt]
Name: 
{agent_name}

Profile:
{profile}

Background:
{background}

Imagine you are {agent_name}, your task is to act/speak as {agent_name} would, keeping in mind {agent_name}'s goal.

\end{lstlisting}
\end{DefaultBox}
\begin{DefaultBox}{Prompt Template for \textsc{Sotopia} Generative Reward Model}
\begin{lstlisting}[language=json, style=agentcode, breakautoindent=false, breakindent=0pt]
## Role
You are an evaluator specializing in Social Intelligence.
Your task is to assess and compare multiple candidate responses within the same interaction context.

## Critical Instructions
You are ONLY evaluating and comparing candidate responses, the output must be a JSON object and nothing else.

## Task
Inputs:
(1) Dialogue context
(2) Assistant's candidate responses
(3) User's reactions to the assistant's candidate responses
(4) {target_agent}'s goal: {target_goal}

## Evaluation Criteria
1. Believability: Naturalness of interactions and consistency with agent traits.
2. Relationship: How interactions affect personal relationships and social standing.
3. Knowledge: Whether new and important information is gained.
4. Secret: Whether confidential or private information is kept or leaked.
5. Social Rules: Compliance with moral rules and laws.
6. Financial and Material Benefits: Positive or negative impact on material or financial status.
7. [most important] Goal: Progress toward and completion of social goals.

### Evaluation Requirements
For EACH candidate response:
1. Provide a detailed critique in the "analysis" field.
2. Do NOT mention sample indices or labels inside the analysis text.
3. Each analysis must focus ONLY on the corresponding response.
4. Assign:
   - A unique integer "rank" (1 = best, larger number = worse)
   - A unique float "score" between 0 and 1
5. No two responses share the same rank or score.


## Output Format (Strict)
Return JSON only. Do not include explanations outside the JSON.
{{
    "0": {{
        "analysis": "Detailed critique of the response.",
        "rank": x,
        "score": 0.xx
    }},
    ...
}}

## Evaluation Request

<Interaction Context>
{messages}
</Interaction Context>

<Candidate Responses>
{samples}
</Candidate Responses

Return a single JSON object. Do not output anything other than the JSON object.
\end{lstlisting}
\end{DefaultBox}

\begin{DefaultBox}{Prompt Template for \textsc{Sotopia} Teacher Policy}
\begin{lstlisting}[language=json, style=agentcode, breakautoindent=false, breakindent=0pt]
Given the dialogue history and the critic feedback, refine the original response accordingly.

<Original Response>
{response}
</Original Response>

<User Reaction>
{user_reaction}
</User Reaction>

<Critic Feedback>
{critic}
</Critic Feedback>

Output only the revised response text. Do not include any explanations, meta-commentary, or prefixes.
\end{lstlisting}
\end{DefaultBox}

\end{document}